\documentclass{article}

\usepackage{PRIMEarxiv}

\usepackage[utf8]{inputenc} 
\usepackage[T1]{fontenc}    
\usepackage[colorlinks=true, urlcolor=blue]{hyperref} 
\usepackage{url}            
\usepackage{booktabs}       
\usepackage{nicefrac}       
\usepackage{microtype}      
\usepackage{lipsum}
\usepackage{graphicx}
\usepackage{enumitem}
\usepackage{orcidlink}
\usepackage{subcaption}
\usepackage{booktabs}
\usepackage{lipsum}
\usepackage{xcolor}
\usepackage{wrapfig}
\usepackage{xspace}
\usepackage{multirow}
\usepackage{balance}
\usepackage{amsmath,amssymb,amsfonts}
\newcommand{\mypara}[1]{\vspace{1mm}\noindent\textbf{#1}}
 
\title{\Large Hierarchical and Step-Layer-Wise Tuning of Attention Specialty for Multi-Instance Synthesis in Diffusion Transformers
}

\author{
  Chunyang Zhang$^\dagger$\\
  School of Systems and Computing\\
  University of New South Wales \\
  Canberra, ACT 2612, Australia\\
  \texttt{chunyang.zhang@unsw.edu.au} \\
   \And
  Zhenhong Sun$^{\dagger\ddagger}$\\
  School of Engineering \\
  Australian National University \\
  Canberra, ACT 2601, Australia\\
  \texttt{zhenhongsun1992@outlook.com} \\
   \AND
  Zhicheng Zhang\\
  School of Business\\
  University of New South Wales\\
  Canberra, ACT 2612, Australia\\
  \texttt{zhicheng.zhang2@unsw.edu.au} \\
       \And
  Junyan Wang\\
  Australian Institute for Machine Learning\\
  University of Adelaide\\
  Adelaide, SA 5371, Australia\\
  \texttt{junyan.wang@adelaide.edu.au} \\
   \And
  Yu Zhang\\
  School of Business\\
  University of New South Wales\\
  Canberra, ACT 2612, Australia\\
  \texttt{m.yuzhang@unsw.edu.au} \\
     \And
  Dong Gong\\
  School of Computer Science and Engineering\\
  University of New South Wales\\
  Syndey, NSW 2052, Australia\\
  \texttt{dong.gong@unsw.edu.au} \\
     \And
  Huadong Mo\thanks{\textbf{Corresponding Authors; $^\dagger$\textbf{Equal Contribution}; $^\ddagger$\textbf{Project Leader.}}}\\
  School of Systems and Computing\\
  University of New South Wales\\
  Canberra, ACT 2612, Australia\\
  \texttt{huadong.mo@unsw.edu.au} \\
       \And
  Daoyi Dong$^*$\\
  School of Computer Science\\
  University of Technology Sydney\\
  Syndey, NSW 2007, Australia\\
  \texttt{daoyi.dong@uts.edu.au}
}

\begin{document}
\maketitle
\vspace{-3em}
\begin{center}
\large
\textbf{Project Page:} \url{https://bitzhangcy.github.io/MIS-DiTs-AST}
\end{center}
\vspace{1em}

\begin{abstract}
Text-to-image (T2I) generation models often struggle with multi-instance synthesis (MIS), where they must accurately depict multiple distinct instances in a single image based on complex prompts detailing individual features. Traditional MIS control methods for UNet architectures like SD v1.5/SDXL fail to adapt to DiT-based models like FLUX and SD v3.5, which rely on integrated attention between image and text tokens rather than text-image cross-attention. To enhance MIS in DiT, we first analyze the mixed attention mechanism in DiT. Our token-wise and layer-wise analysis of attention maps reveals a hierarchical response structure: instance tokens dominate early layers, background tokens in middle layers, and attribute tokens in later layers. Building on this observation, we propose a training-free approach for enhancing MIS in DiT-based models with hierarchical and step-layer-wise attention specialty tuning (AST). AST amplifies key regions while suppressing irrelevant areas in distinct attention maps across layers and steps, guided by the hierarchical structure. This optimizes multimodal interactions by hierarchically decoupling the complex prompts with instance-based sketches. We evaluate our approach using upgraded sketch-based layouts for the T2I-CompBench and customized complex scenes. 
Both quantitative and qualitative results confirm our method enhances complex layout generation, ensuring precise instance placement and attribute representation in MIS.


\end{abstract}    

\section{Introduction}
\label{sec:intro}
\begin{wrapfigure}{r}{0.50\linewidth}
    \centering
    \includegraphics[width=\linewidth]{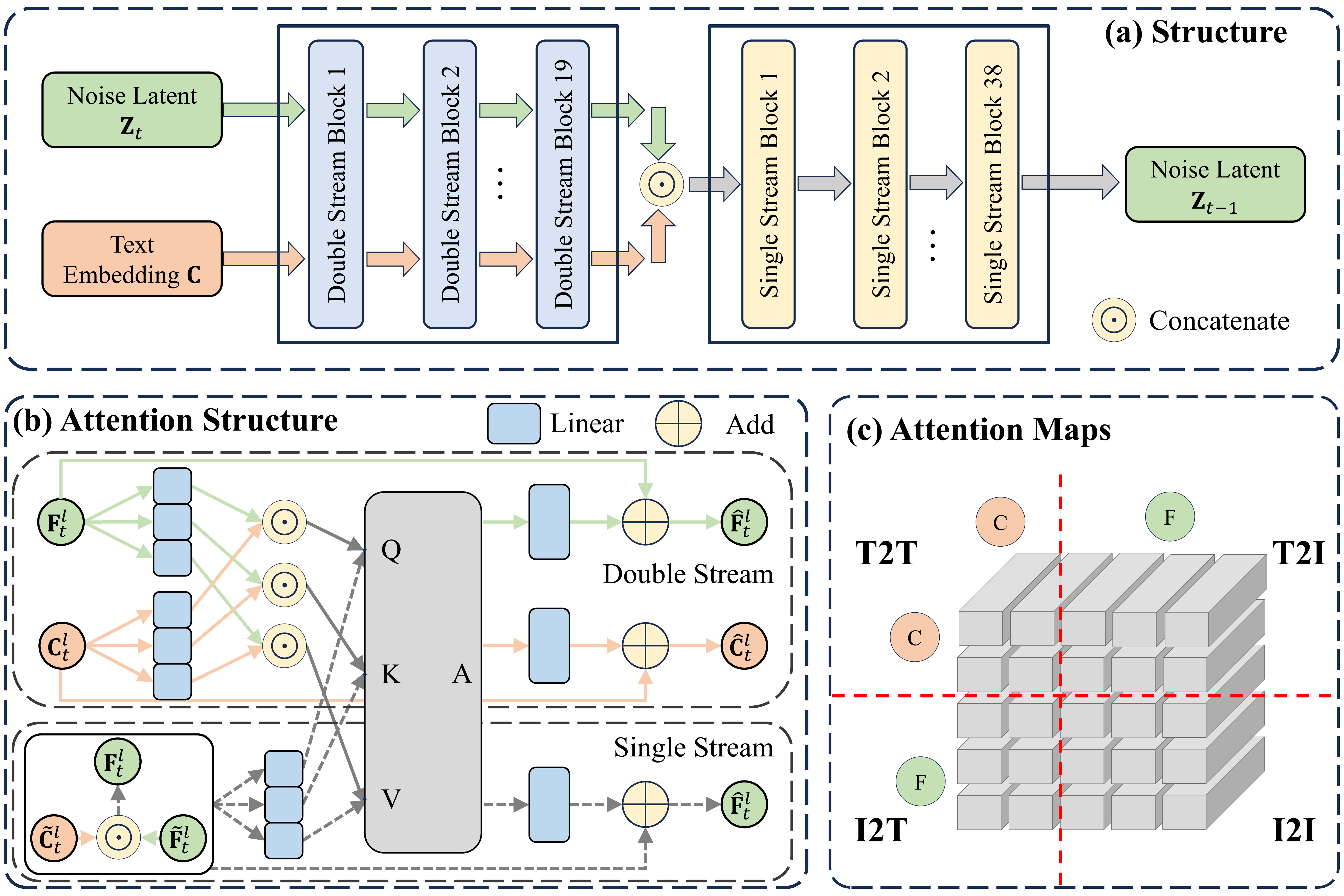}
    \caption{Schematic architecture of the DiT-based models (a) with double/single-stream attentions (b).  The attention map in (c) reveals four distinct interaction patterns: text-to-text, text-to-image, image-to-text, and image-to-image, enabling deep consistent fusion between text and tokens.}
    \label{fig:flux}
\end{wrapfigure}

Multi-instance synthesis (MIS) is a key challenge in text-to-image (T2I) generation, essential for enabling generative models' real-world applications 
\cite{kim2023dense, li2022grounded, li2023gligen, zheng2023layoutdiffusion, yang2023reco, xie2023boxdiff}. It focuses on accurately representing multiple distinct instances within a single image, responding precisely to complex prompts that specify individual features for each instance. 
Leveraging UNet-based diffusion models 
(e.g., SD v1.5~\cite{rombach2022high} and SDXL~\cite{podell2023sdxl}), current generative models~\cite{kim2023dense, li2022grounded, li2023gligen, lian2023llm, zheng2023layoutdiffusion, yang2023reco, xie2023boxdiff, zhou2024migc, sun2024eggen, phung2024grounded, chen2024training, yang2024mastering, singh2024smartmask, lin2024ctrl, wang2024instancediffusion} typically adopt a divide-and-conquer approach to enhance MIS by processing instances and attributes separately, employing either fine-tuning or training-free methods.
Fine-tuning leverages task-specific data to enhance synthesis quality but demands extensive resources for data collection and training, while training-free methods utilize inherent model capabilities offering flexible adaptability, and low computational cost, making them a desirable approach for practical MIS applications.
Recently, Diffusion Transformer (DiT)-based models such as FLUX~\cite{blackforestlabs_flux_2024} and SD v3.5~\cite{esser2024scaling} have demonstrated significant advances in text-image consistency and high-quality image generation. Transitioning from traditional UNet architectures with cross-attention mechanisms to more sophisticated DiT architectures~\cite{peebles2023scalable} utilizing combined attention mechanisms has notably improved complex prompt handling and image representation. However, these enhancements have not explicitly targeted improvements for MIS, especially in scenarios involving intricate prompts and complex spatial layouts.
Moreover, this transition further complicates the application of existing training-free approaches, which were initially designed around UNet architectures. Consequently, the emerging shift also highlights a pressing need to investigate attention mechanisms tailored specifically for DiT architectures that can facilitate effective training-free solutions for MIS tasks.


The DiT architecture replaces traditional self- and cross-attention with a unified attention, and injects text embeddings at each inference step, enabling consistent fusion of textual and visual information.
To explore this mechanism, we analyze specific attention dynamics, focusing on text-to-text (T2T), image-to-image (I2I), and image-to-text (I2T) interactions. Attention map analysis reveals hierarchical responses: instance tokens dominate early layers, background tokens in middle layers, and attribute tokens in later layers, with critical integration mostly in early inference steps.
A step-layer-wise token-exchange experiment further examines the DiT's text embeddings interaction by swapping background, attribute, and instance tokens at specific layers and steps, which confirms the above hierarchical attention responses.
Collectively, these analyses clarify the specific roles of various tokens and provide valuable insights, forming a strong foundation for developing controlled MIS.

Building on the outlined observations, we propose a training-free approach for enhancing MIS in DiT models with hierarchical and step-layer-wise \textbf{A}ttention \textbf{S}pecialty \textbf{T}uning (\textbf{AST}), focusing on the crucial role of attention maps. 
Specifically, AST employs a unified scalable module with specialized masks to improve the attention distribution for distinct T2T, I2I, and I2T attentions. It enhances synthesis by selectively amplifying key regions and suppressing irrelevant areas in attention maps, improving alignment between specific textual or visual regions for enhanced details.
\textbf{H}ierarchical and \textbf{S}tep-\textbf{L}ayer-\textbf{W}ise (\textbf{HSLW}) module refines alignment by modulating attention across layers and steps, designating specific layers for attributes, instances, and background hierarchically, which optimizes multimodal interactions within T2T, I2I, and I2T modules by hierarchically decoupling the complex prompts with instance-based sketches.
Meanwhile, we introduce an upgrade patch with sketch-based layouts for the T2I-CompBench~\cite{kaiyi2024t2ibench} and customized complex scenes to evaluate the performance, where our method achieves accurate instance placement and attribute representation for complex layouts in the MIS. 
The contributions of our work are outlined as follows:
\begin{itemize}[leftmargin=*, noitemsep, nolistsep]
    \item[$\bullet$] We explore the mixed attention mechanism of the DiT model and design the token exchange experiment, revealing the influence of hierarchical attention responses.
    \item[$\bullet$] HSLW-based AST highlights key regions and suppresses irrelevant areas across layers and steps, optimizing multimodal interactions via instance-based sketches.
    \item[$\bullet$] Benefiting from AST, our model extends the DiTs with an improved fusion of textual and visual representations, realizing high-quality MIS.
\end{itemize} 
\section{Related Work}
\label{sec:relate}

\mypara{T2I Synthesis.}
Diffusion models~\cite{sohl2015deep, song2019generative, ho2020denoising, song2020denoising, croitoru2023diffusion} have shown remarkable performance in generating high-quality images for T2I synthesis and editing~\cite{nichol2021glide, saharia2022photorealistic}. 
Latent Diffusion Models (LDMs)~\cite{rombach2022high} further enhance both diversity and realism while maintaining low computational complexity, utilizing the Variational AutoEncoder~\cite{kingma2013auto} and UNet architecture~\cite{ronneberger2015unet} with a cross-attention mechanism. 
Recently, unified transformer-based models~\cite{peebles2023scalable, chen2023pixart, esser2024scaling, ma2024sit, blackforestlabs_flux_2024} represented by DiT~\cite{peebles2023scalable}, combined with the rectified flow technique~\cite{liu2022flow,esser2024scaling}, are increasingly demonstrating their superiority compared with UNet architecture. 
Notably, DiT models have achieved new state-of-the-art performance in image synthesis.
However, the DiT-based models face challenges in controllable MIS when dealing with complex prompts and specific layouts.

\mypara{Multi-Instance Synthesis.}
MIS focuses on accurately generating images containing multiple distinct instances that align with specific attributes and layouts~\cite{wu2023harnessing, kim2023dense, li2022grounded, li2023gligen, lian2023llm, zheng2023layoutdiffusion, yang2023reco, xie2023boxdiff, zhou2024migc, sun2024eggen, phung2024grounded, chen2024training, yang2024mastering, singh2024smartmask, lin2024ctrl, wang2024instancediffusion}.
Based on the type of layout input, methods can be divided into bounding box-based~\cite{li2022grounded, li2023gligen, lian2023llm, zheng2023layoutdiffusion, yang2023reco, zhou2024migc, sun2024eggen, phung2024grounded, xie2023boxdiff, wu2023harnessing, chen2024training, yang2024mastering} and sketch-based~\cite{kim2023dense, singh2024smartmask, lin2024ctrl, wang2024instancediffusion,sun2024t} approaches. Also, these methods can be further classified into fine-tuning methods that leverage formatted data~\cite{li2022grounded, li2023gligen, lian2023llm, zheng2023layoutdiffusion, yang2023reco, zhou2024migc, sun2024eggen, phung2024grounded, singh2024smartmask, lin2024ctrl, wang2024instancediffusion} and training-free methods that use cross-attention mechanisms directly~\cite{xie2023boxdiff, wu2023harnessing, chen2024training, yang2024mastering, kim2023dense, xin2025dymo}.
Adapting fine-tuning on the DiT-based models could improve MIS, but the high computational costs make a training-free method more efficient.
Unlike bounding box layouts, sketch layouts offer flexible shapes and semantic details, making them accessible to a wider non-artistic audience.
Therefore, we aim to develop a training-free method using sketch layouts on the DiT-based models for effective MIS.

\mypara{Attention Tuning Synthesis.} 
In the UNet architecture with the cross attention mechanism, works~\cite{hertz2022prompt, feng2022training, ruiz2023dreambooth, zhou2023maskdiffusion, hong2023improving, wang2024towards} use tuning strategies to refine attention maps, guiding the generation process and enhancing the model's accuracy in following prompts. 
For example, Prompt-to-Prompt~\cite{hertz2022prompt} examines attention maps and introduces training-free attention tuning to improve alignment between textual prompts and specified instances. 
P+ \cite{voynov2023p+} manipulates prompt token replacement across layers to reveal each layer's influence on token behavior, while the HCP model~\cite{wang2024towards} illustrates how tokens react differently at various inference steps. 
Meanwhile, Dense Diffusion~\cite{kim2023dense} uses sketch layouts to target specific regions, adjusting self- and cross-attention maps during denoising to enhance image generation.
However, as the focus shifts to DiT-based models, traditional methods effective in the UNet architecture face challenges, which underscores the need to explore new tuning mechanisms within DiT to develop an efficient training-free approach.

\begin{figure*}[t]
    \centering
    \includegraphics[width=1\linewidth]{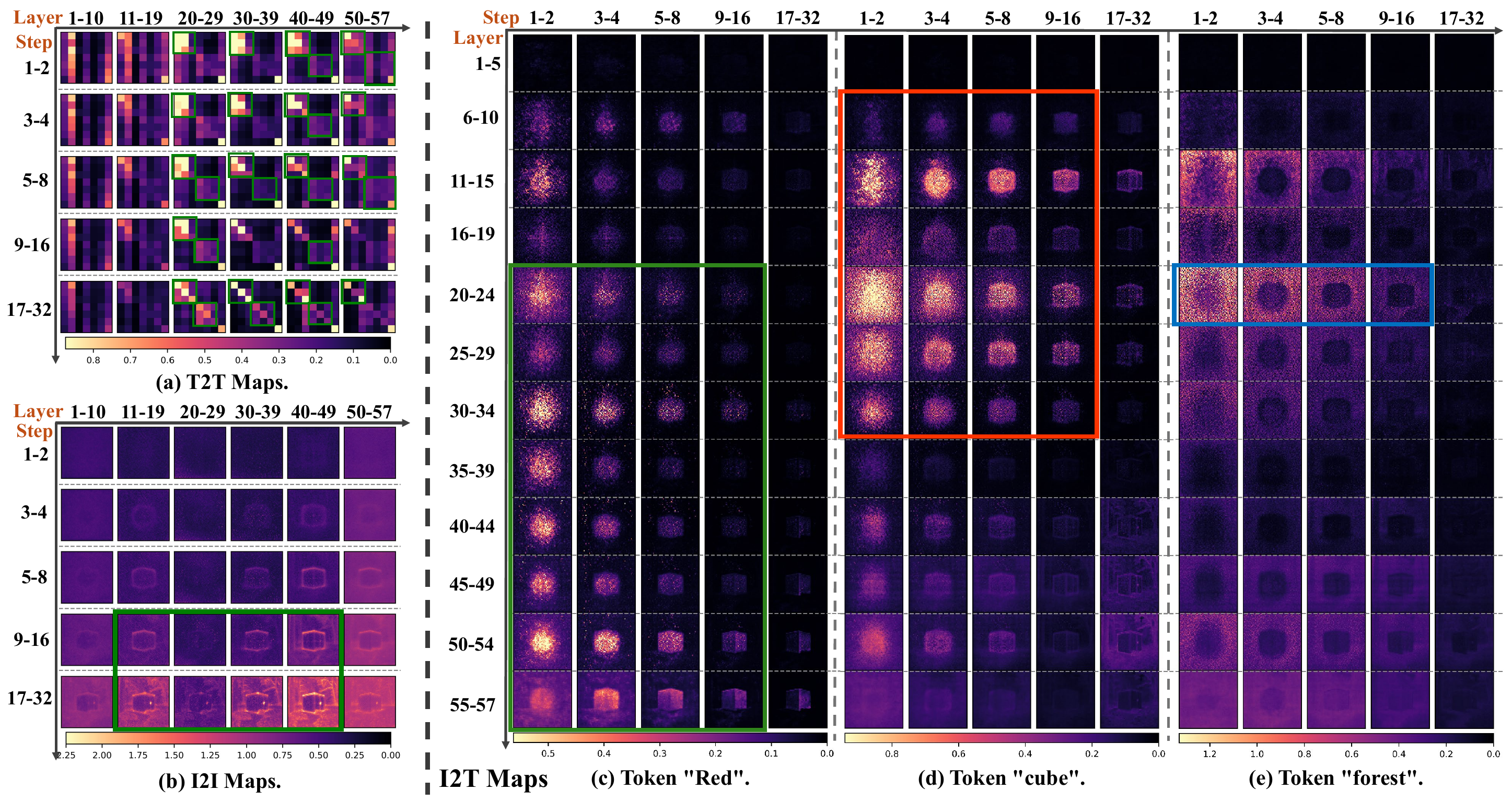}
    \caption{Attention map averages for the prompt ``Red cube in a forest". 
    (a) T2T maps show strong intra-segment interactions within valid tokens of ``Red cube" (first 3 tokens) and ``in a forest" (last 4 tokens).
    (b) I2I maps (the maximum value of each point relative to all other points) reveal growing interaction between ``Red cube" and ``a forest" segments over steps.
    (c/d/e)  I2T attention maps on individual tokens highlight that instance tokens dominate early layers, background tokens in the middle, and color tokens later, with most information integrated in the first half of steps. T2I maps showing lower scores with weak impact and more examples are present in the \textbf{Appendix.\ref{app-section:8}}.}
    \label{fig:self}
\end{figure*}    

\section{Proposed Approach}
\label{sec:method}
In this section, our approach begins with an examination of the DiT-based model, emphasizing the role of new attention mechanisms. Following this analysis, we design a step-layer-wise token-exchange experiment to investigate prompt token replacement across layers. 
Subsequently, we propose a training-free HSLW module of AST for MIS with DiT-based models, followed by detailed explanations.

\subsection{Preliminary}
\label{subsec:preliminary}
Unlike the traditional UNet structure, the DiT architecture~\cite{peebles2023scalable} eliminates spatial downsampling and upsampling to sustain a consistent higher spatial dimension. 
It substitutes the original self- and cross-attention mechanisms with a unified attention mechanism, simplifying the architecture, enhancing efficiency, and improving generalization effectiveness~\cite{peebles2023scalable, esser2024scaling, blackforestlabs_flux_2024}. 
Instead of injecting text embeddings via cross-attention at each layer, the DiT-based model introduces them once at the start of each inference step, allowing them to merge with the noise latent for deeper and spatially consistent text-visual fusion, as shown in Figure~\ref{fig:flux}.
The DiT-based model employs double stream blocks that initially combine latent and text embeddings through separate streams, followed by single stream blocks that unify them into a single stream, as illustrated in Figure~\ref{fig:flux} (a).
Despite the structural differences of the two streams, they adhere to a new attention mechanism, with the intermediate output feature $\mathbf{\check{F}}^l_t$ defined by:
\begin{equation} 
\mathbf{\check{F}}^l_t = \mathbf{A}^l_t \mathbf{V}^l_t = \text{softmax}\left( \frac{\mathbf{Q}^l_t {\mathbf{K}^l_t}^T}{\sqrt{d^l_c+d^l_z}} \right) \mathbf{V}^l_t, 
\end{equation}
where $\mathbf{Q}^l_t$, $\mathbf{K}^l_t$ and $\mathbf{V}^l_t$ reside in the dimension of $ \mathbb{R}^{(d_c+hw)\times d^l}$, transformed by the linear layers from the latent features and text embeddings, as depicted in Figure~\ref{fig:flux} (b). Here, $d_c$ represents the max tokens of text embeddings, $hw$ is the spatial dimensions, and $d^l$ is the embedding dimension of linear layers.
Consequently, the attention maps $\mathbf{A}^l_t\in \mathbb{R}^{(d_c+hw)\times (d_c+hw)}$ are divided into four parts (Figure~\ref{fig:flux} (c)), covering T2T, T2I, I2T and I2I modules.
This architecture functions as an equivalent blend of the original self- and cross-attention, improving text-visual fusion.

\subsection{Attention Analysis of DiT Model}
\label{subsec:analysis}

\mypara{Hierarchical Attention Responses. }
Following the analytical approaches~\cite{hertz2022prompt,wang2024towards} based on the UNet architecture, we utilize the prompt ``Red cube in a forest" to examine the functionality of four regions in the attention maps within the DiT-based  model. For simplicity, we introduce a prompt that includes an instance with an attribute alongside a background element, allowing an intuitive investigation. 
The visualization maps are categorized by types as shown in Figure~\ref{fig:self}: self-attention maps (T2T and I2I), and cross-attention-based I2T maps. Notably, the T2I attention shows lower attention scores relative to others, indicating a weak impact on generation, which is provided in \textbf{Appendix.\ref{app-section:8}}. We have the following observation.

\mypara{1) Text-to-Text:} This region enhances intra-segment correlations while minimizing inter-segment interference. Specifically, ``Red cube" (3 tokens) and ``in a forest"(4 tokens) show high internal attention scores, indicating strong token interactions within each sub-prompt, supporting instance-specific responses, and decreasing cross-attribute coupling.

\mypara{2) Image-to-Image.} This region controls the interactions of instances, allowing elements like the ``cube" and ``forest" to emerge through denoising gradually. Early stages focus on the instance’s internal structure, while later stages increase interaction with the background, suggesting that interventions in the instance are most effective in early stages.

\mypara{3) Image-to-Text.} This region distinguishes instances with attributes and background activating at different but early stages: the attribute ``Red" in layers 20-57,  the instance ``cube" in layers 6-34, and the background ``forest" in layers 20-24. This step-layer-wise distribution suggests a precise scope for attention manipulation, particularly in layers where background and instance overlap, allowing for targeted adjustments to realize high efficiency.

In summary, this analysis reveals that HSLW attention specificity in I2T enhance text responsiveness in the text-to-query region while defining semantic and structural features in the image-to-query region. This enables DiT-based model to generate coherent images, integrating foreground instances with attributes into the background and providing key priors for attention manipulation.

\begin{figure*}[t]
    \centering
    \includegraphics[width=0.93\textwidth]{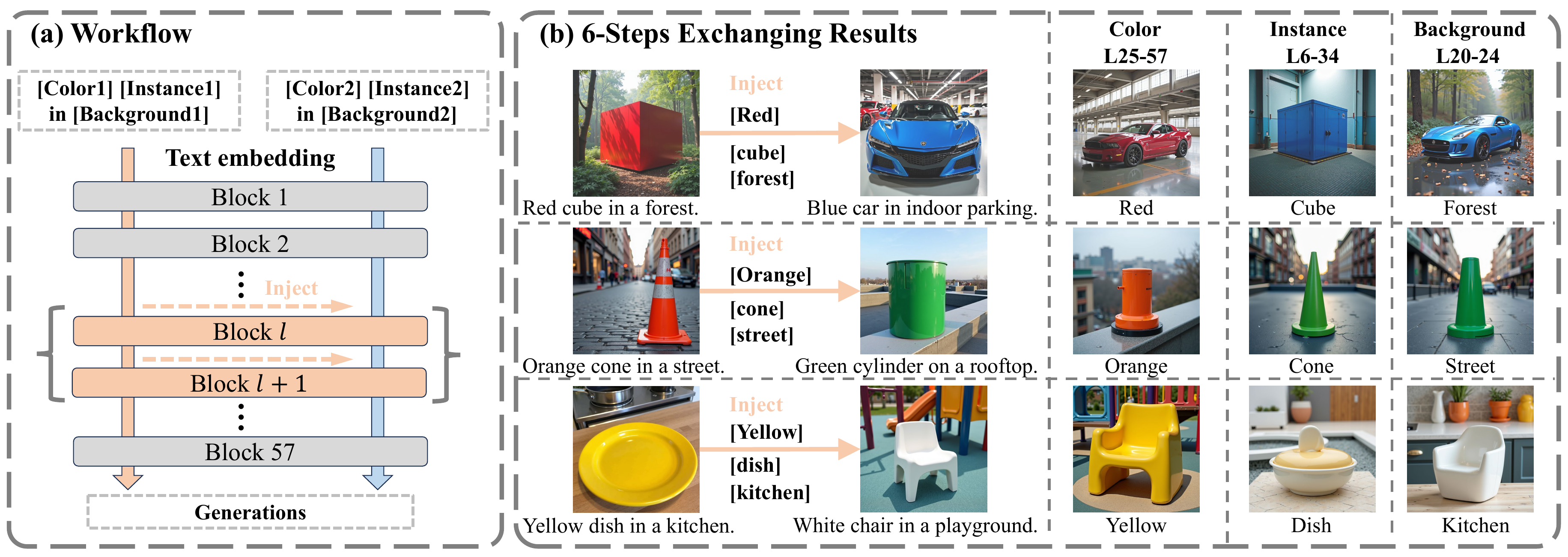}
    \caption{Step-Layer-Wise Token Exchange involved injecting specific tokens between two prompts formatted as ``[color] [Instance] in [Background]" (a). First 6-step exchanging results show that specific token concept can be replaced in certain layers (e.g., L25-27) with other concepts preserving, aligning with the hierarchical attention responses in Figure~\ref{fig:self}. Step-wise visualization is detailed in \textbf{Appendix.\ref{app-section:9}}.}
    \label{fig:ex-trial}
\end{figure*}

\mypara{Step-Layer-Wise Token Exchanging. }
As mentioned in Sec.~\ref{subsec:preliminary}, the DiT model introduces text embeddings once at the start of each inference, allowing them to flow alongside the noise latent.
This structure supports enhanced integration of textual and visual information at a uniform spatial scale, although its underlying role remains unexamined.
Inspired by the P+ model~\cite{voynov2023p+}, which manipulates prompt token replacement across layers, we propose a step-layer-wise token exchanging. The exchange involved injecting specific tokens at different layers between two prompts formatted as “[color] [Instance] in [Background]”, to observe the evolution of text embeddings during diffusion (Figure~\ref{fig:ex-trial} (a)). The resulting images in Figure~\ref{fig:ex-trial} (b) demonstrate that selective component replacements, such as color, instance, and background, operate on different layers, supporting our earlier findings on hierarchical attention responses and step-layer-wise specificity in I2T module. For instance, elements such as ``red", ``cube", and ``forest" prominently emerge by step 6 in a harmonious style with other concepts preserving.  These findings provide a foundation for developing an attention tuning on the DiT model for controllable MIS with complex prompts and layouts.

\subsection{Hierarchical Tuning}
\label{subsec:overview}
\begin{figure*}[t]
    \centering
    \includegraphics[width=0.95\textwidth]{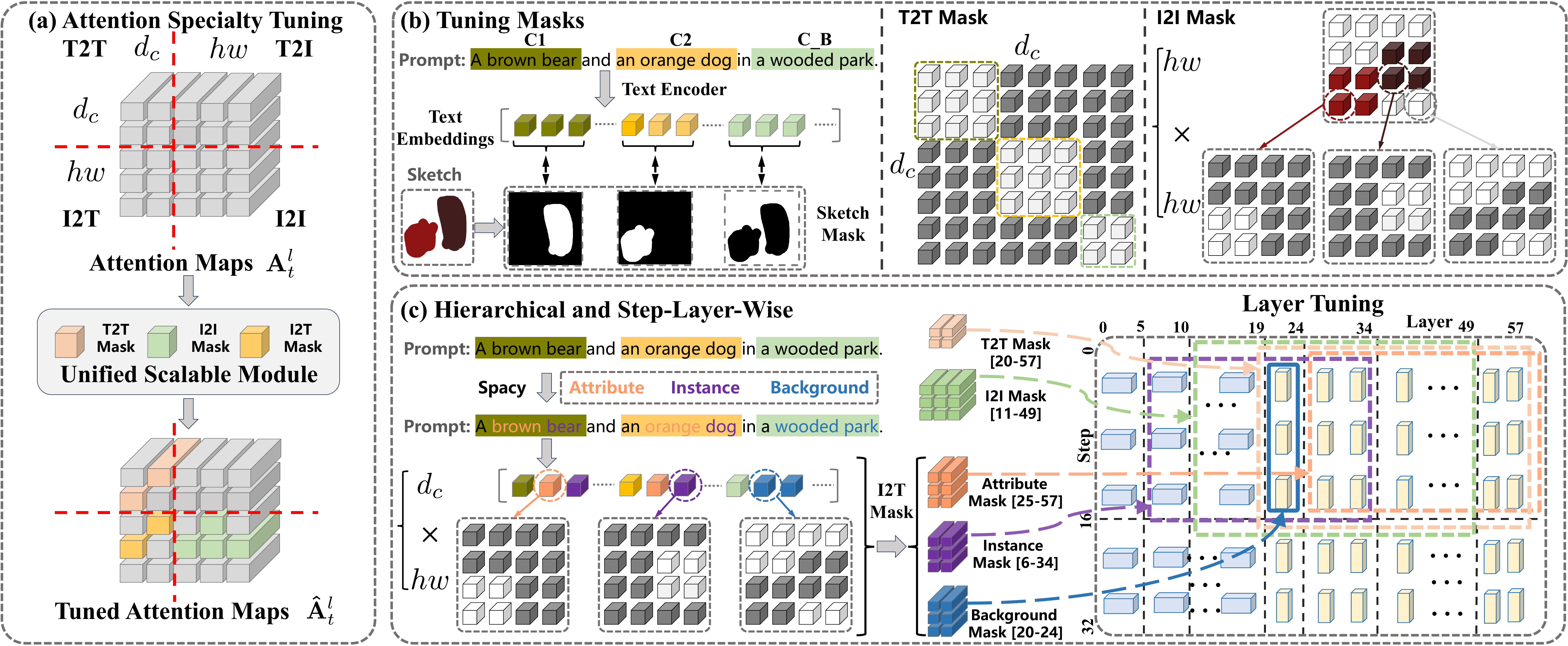}
    \caption{Overview of the hierarchical tuning strategy for DiT models. (a) Attention specialty tuning enhances meaningful regions while suppressing non-meaningful ones via a unified scalable module with distinct tuning masks.
    (b) Tuning masks for T2T and I2I masks are directly built from prompt segments and sketches. 
    (c) Hierarchical and step-layer-wise module refines alignment by adjusting each component’s impact across specific layers and steps, assigning layers for attributes, instances, and background tokens within first 16 steps.}
    \label{fig:overview}
    \vspace{-6pt}
\end{figure*}

To address the challenges of the attention mechanism discussed in Section \ref{subsec:analysis}, we propose a training-free hierarchical tuning strategy, which leverages the strengths of the pre-trained DiT-based model with hierarchical and step-layer-wise AST, incorporating: textual prompts $\mathbf{c}$ and corresponding sketch images $\mathbf{S}$, as shown in Figure~\ref{fig:overview} . 



\mypara{Attention Specialty Tuning. }
The essence of attention tuning in the cross-attention mechanism is to enhance the values of meaningful regions in the attention maps while attenuating the values of weak or non-meaningful regions~\cite{hertz2022prompt,voynov2023p+,kim2023dense,wang2024towards}. However, the DiT-based models' attention mechanism comprises self-focused T2T and I2I, as well as cross-focused I2T and T2I attention maps, each exhibiting unique characteristics that influence generation outcomes. 
Therefore, effective attention tuning of DiT-based models requires accounting for these variations to enhance instance accuracy and detail in MIS. 
Thus, we propose a plug-in strategy for the mixed attention mechanism shown in Figure~\ref{fig:overview} (b), named AST.

We design a unified scalable module to process different attention types consistently. Adjustments are focused on mask specification to enhance or attenuate specific regions within each attention category. This unified structure streamlines processing across attention types, improving model efficiency, scalability, and adaptability in highlighting meaningful regions. The process of tuning original attention map \(\mathbf{A}^l_t\) (after softmax) is specified as follows:
\begin{align}
\mathbf{\hat{A}}^l_t=f_{norm}({\mathbf{A}^l_t}\odot\exp(\beta_t\cdot\mathbf{G}\odot(\mathbf{M}-\mathbf{A}^l_t)),
\label{eq:atten}
\end{align}
where \(\mathbf{\hat{A}}^l_t\) represents the tuned attention maps at layer \(l\) and time step \(t\), and \(\odot\) denotes element-wise multiplication (Comparison before softmax is in \textbf{Appendix.\ref{app-section:7}}).
\(f_{norm}(\mathbf{B})={\mathbf{B}}/{\sum_i B_i} \) is the weight normalization to form a distribution summing to 1 along the embedding dimension. Here, \(\beta_t=\lambda (t/T)^4\) is a scaling factor that adjusts the sensitivity of the enhancement with a hyper-parameter $\lambda$ and maximum \(T\) steps. \(\mathbf{G}\) is a sensitivity matrix to modulate region-specific adjustments (following Dense Diffusion~\cite{kim2023dense}, the specific formula is provided in \textbf{Appendix.\ref{app-section:3}}), with values determined by the proportion of the sketch area relative to the overall image area—smaller objects approach 1, while larger objects approach 0.
Notably, \(\mathbf{M}\) is the mask matrix specific to different attention types, incorporating conditions corresponding to prompts $\mathbf{c}=\{\mathbf{c}_1, \mathbf{c}_2, ..., \mathbf{c}_N\}$ and sketch images $\mathbf{S}=\{\mathbf{S}_1, \mathbf{S}_2, ..., \mathbf{S}_N\}$ ($N$ is the number of sub-prompts), and guiding adjustments to attention values. Relying on the characteristics of the attention maps, each attention employs distinct constructions:

1) \textbf{T2T Mask}: In the attention maps, the T2T region (sized 
$d_c \times d_c$) represents the correlation between elements within the text embeddings. To strengthen the coupling among sub-prompt elements, as concluded in Section~\ref{subsec:analysis}, we apply a binary mask matrix within sub-prompts, as shown in Figure~\ref{fig:overview} (b), and expressed by:
\begin{align}
\mathbf{M}_{\text{T2T}}(i, j) = 1 \quad \text{if } i, j \in \mathbf{c}_k, \quad k \in \{1, \dots, N\}.
\end{align}

2) \textbf{I2I Mask}: I2I region ($hw \times hw$) captures correlations between each element and all others, promoting internal coherence across image features. Therefore, the mask should represent the alignment of visual elements within a sub-prompt instance (Figure~\ref{fig:overview} (b)), reinforcing the associations within the instance, and is calculated as:
\begin{align}
\mathbf{M}_{\text{I2I}}(i, j) = 1 \quad \text{if } i, j \in \mathbf{S}_k.
\end{align}

3) \textbf{I2T Mask}: I2T region ($hw \times d_c$) represents the alignment between image and text embeddings. For each text embedding corresponding to sub-prompts, attention values should be enhanced at positions that align with the corresponding sketch locations in latent space (Figure~\ref{fig:overview} (c)), reinforcing cross-modal correspondence between visual features and textual elements within the prompts, defined by:
\begin{align}
\mathbf{M}_{\text{I2T}}(i, j) = 1 \quad \text{if } i \in \mathbf{S}_k \text{ and } j \in \mathbf{c}_k.
\end{align}

\mypara{Hierarchical and Step-Layer-Wise Module. }
HSLW module of different attention types effectively handles the coupled alignment of textual information and visual layout across different instances, especially for T2T and I2I. However, the I2T module, which ultimately governs multimodal interaction, lacks sufficient details in tuning. As discussed in Section~\ref{subsec:analysis}, the background, instance, and attribute are coupled in the spatial dimension but have distinct regions of influence (Figure~\ref{fig:self} (c)). Thus, a step-layer-wise modulation for I2T focusing on background, instance, and attribute is introduced to better respond to textual information for each instance.

Specifically, for different sub-prompts, we employ an NLP library (e.g., SpaCy) to distinguish between adjectives (attributes) and nouns (instances) words. Adjectives and instance words classified as background words are marked by their respective index positions in the text embedding. 
Corresponding to the hierarchical attention responses, an instance tuning is applied to the earlier layers (layers 6 to 34), a background tuning is incorporated into the intermediate layers (layers 20 to 24) while a attribute tuning is applied to the later layers of the network (layers 25 to 57), as shown in Figure~\ref{fig:overview} (c). This hierarchical allocation of I2T layers helps achieve a more precise alignment between textual and visual information in complex prompts, improving the response of MIS. Moreover, distinct steps and layers are applied for T2T and I2I to analyze the self-focused components, as illustrated in Figure~\ref{fig:self}. T2T attention should be emphasized in layers 20 to 57 to enhance intra-segment token interactions and avoid attribute and instance leakage due to strong correlations in the first 16 steps. Similarly, I2I attention should be refined in the early layers (11 to 49) to better capture the layout relationships between background and foreground instances. 

\section{Experiments}
\label{sec:exp}

\begin{table}[!ht]
\centering
\caption{Evaluation results on T2I-CompBench with upgraded sketches (\textbf{Appendix.\ref{app-section:5}}) and image quality assessment on customized complex cases (\textbf{Appendix.\ref{app-section:6}}) with study scores (1-5) and ImageReward~\cite{xu2023imagereward}. ImageReward shows that fine-tuning the base model degrades image quality usually, whereas our method results in a slight drop compared with others.
\textit{$^\#$} denotes results from T2I-CompBench, while others are re-evaluated. Inference time is measured with identical 50 denoising steps. ``Ours-v1" indicates FLUX with I2T mask, ``Ours-v2" indicates FLUX with I2T+I2I masks and ``Ours-v3" indicates FLUX with I2T+T2T mask.
}\label{tab:main-results}
\setlength{\tabcolsep}{3pt} 
    \begin{tabular}{c|ccc|cc|c}
    \toprule[1pt]
    Datasets & \multicolumn{3}{c|}{T2I-CompBench} & \multicolumn{2}{c|}{Customized Cases} &\\
    \midrule[1pt]
    \multirow{2}{*}{Metrics} & Color  & 2D Spatial & Complex  & User & Image& Infer \\
     & B-VQA $\uparrow$
    & UniDet $\uparrow$
    & 3 in 1 $\uparrow$ & Study$\uparrow$ & Reward$\uparrow$ & s/it 
    \\
    \midrule[1pt]
    SD v1.5\textit{$^\#$}~\cite{rombach2022high} & 0.3765 & 0.1246 & 0.3080 &  1.74  & -0.448 & 7\\
    SD v2\textit{$^\#$}~\cite{rombach2022high} & 0.5065 & 0.1342 & 0.3386&  2.21& -0.1482 &6 \\
    SDXL\textit{$^\#$}~\cite{podell2023sdxl}& 0.6369 & 0.2032 & 0.3237&  2.76&0.48 & 14 \\
    Pixart\textit{$^\#$}~\cite{chen2023pixart}& 0.6690 & 0.2064 & 0.3433&  3.01& 0.8584 & 14 \\
    SD v3.5~\cite{esser2024scaling} & 0.7484 & 0.3231 & 0.5334 & 3.29 & 1.4592 &62\\
    FLUX-dev~\cite{blackforestlabs_flux_2024} & 0.7398 & 0.2589 & 0.5081 & 3.26 &1.2371 & 23\\
    \midrule[1pt]
    DenseDiff~\cite{kim2023dense}& 0.4316 & 0.2142 & 0.3563 &  2.17& -0.352 &12\\
    BoxDiff~\cite{xie2023boxdiff} & 0.5146 & 0.2582 & 0.4044 &1.76 & -0.3402&10\\
    Atten-Refoc~\cite{phung2024grounded} & 0.4442 & 0.4503 & 0.3966 & 2.21&-0.0564&37\\
    HiCO~\cite{cheng2024hico} & 0.5500 & 0.3181 & 0.4274 &2.42 &-0.1865 &10\\
    R\&B~\cite{xiao2023r} & 0.5069 & 0.3841 & 0.4204 & 1.86& -0.6838&40\\
    GrounDiT~\cite{lee2024groundit} & 0.5997 & 0.4409 & 0.5031 & 2.71&0.5521 &130\\
    MIGC~\cite{zhou2024migc} & 0.4974 & 0.4628 & 0.4974 &2.48&-0.184&9\\
    Instance~\cite{wang2024instancediffusion}& 0.6583 &0.4623 & 0.5059 &3.14 &0.1888&14\\
    RPG~\cite{yang2024mastering} & 0.4724 & 0.2517 & 0.3850 &3.01 &0.0277&20\\
    \midrule[1pt]
    Ours+SD v3.5  & 0.7531 & 0.4188 & 0.5567 & 3.63 &1.3257& 94 \\
    Ours+FLUX  & 0.7822 & 0.5004 & 0.5717 &4.03&0.9213 &29\\
    \midrule[1pt]
    Ours-v1  & 0.7459 & 0.4383 & 0.5456 & 3.41&0.8943&29\\
    Ours-v2  & 0.7854 & 0.4569 & 0.5654 &3.61 &0.7788&29\\
    Ours-v3  & 0.7853 & 0.4403 & 0.5619 &3.76 & 0.8338&29\\
    \bottomrule[1pt]
    \end{tabular}
    \vspace{-6pt}
\end{table}

\subsection{Implementation Details}
\label{subsec:imp}

\begin{figure*}[!ht]
    \centering
    \includegraphics[width=0.95\textwidth]{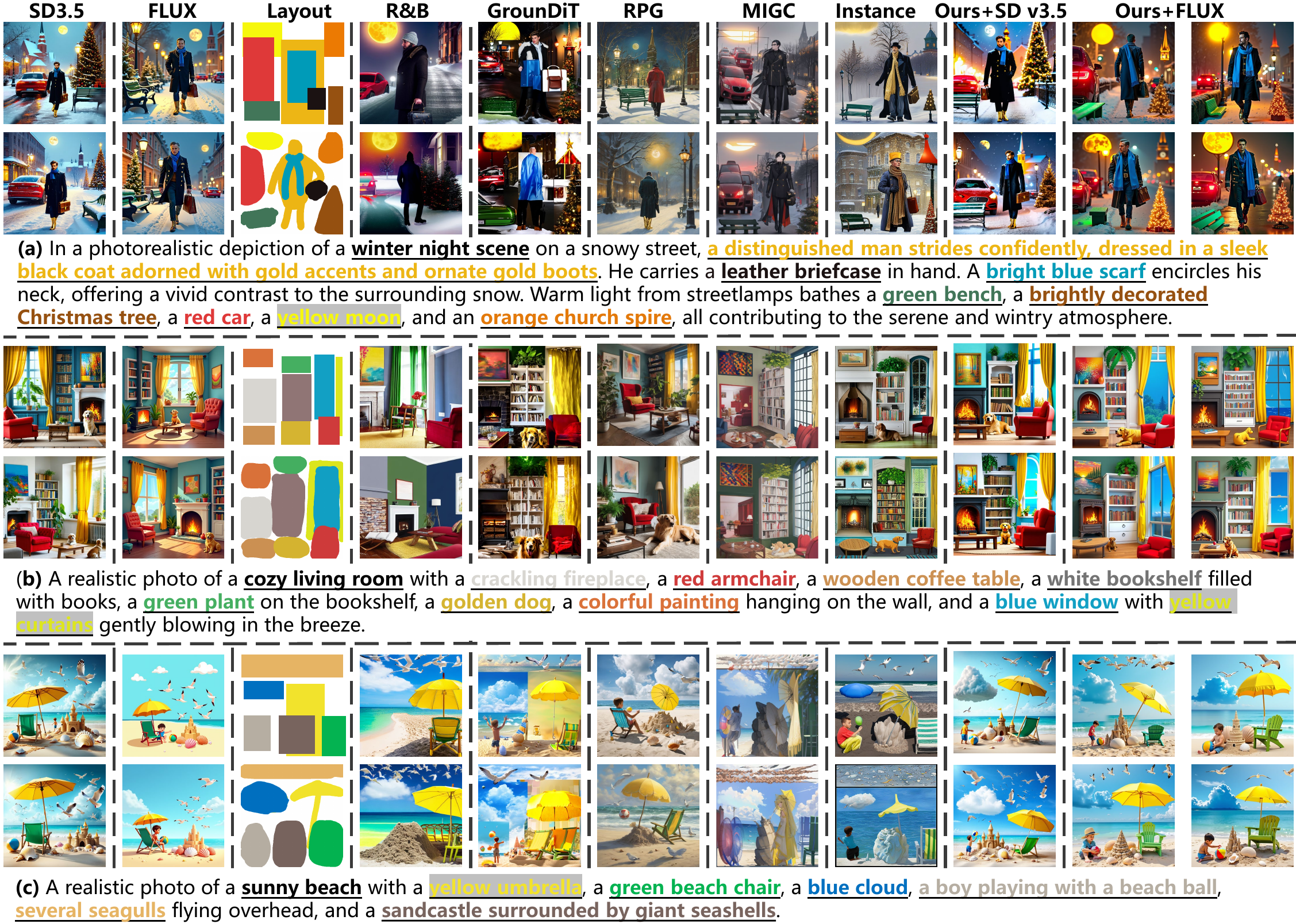}
    \caption{Visualizations on customized complex cases (More examples are in \textbf{Appendix.\ref{app-section:6}}). Bounding box layouts are for MIGC~\cite{zhou2024migc} Instance~\cite{wang2024instancediffusion}, R\&B~\cite{xiao2023r}, and GroundDiT~\cite{lee2024groundit}, and the layout of RPG~\cite{yang2024mastering} is generated by the GPT-4o~\cite{openai2024gpt4technicalreport}, while sketch layouts are for ours. Our approach extends DiT models (SD v3.5 and FLUX) with better multi-instance alignment with prompts and sketch-based layouts. The underlined prompts within the prompts are the sub-prompts for instances, with colors matching those in the sketch.}
    \label{fig:main}
    \vspace{-6pt}
\end{figure*}

\mypara{Baselines and Settings.}  In our experiments, we utilize the FLUX-dev model~\cite{blackforestlabs_flux_2024} as both the analysis and evaluation model, and SD v3.5~\cite{esser2024scaling} as the evaluation model. The hyperparameters $\lambda$ in Eq.~(\ref{eq:atten}) are set to $\lambda_{cross} = 5.0$ and $\lambda_{self} = 3.5$. During inference, we apply the default Euler Discrete Scheduler~\cite{esser2024scaling} with 32 steps and a guidance scale of 7 at a resolution of $1024\times1024$. Inferences are conducted on 1 Nvidia V100 GPU with 32 GB of memory, while analysis experiments are performed on 1 A100 GPU with 80 GB of memory to accommodate the storage of attention maps sized 4608×4608.
To evaluate our method, we compare it against several baseline models, including
SD v1.5~\cite{rombach2022high}, SD v2\cite{rombach2022high}, SDXL~\cite{podell2023sdxl},
and Pixart~\cite{podell2023sdxl}, and related training-free/based methods, including
DenseDiff~\cite{kim2023dense}, BoxDiff~\cite{xie2023boxdiff}, 
Atten-Refoc~\cite{phung2024grounded}, HiCO~\cite{cheng2024hico}, 
R\&B~\cite{xiao2023r},
GrounDiT~\cite{lee2024groundit},
MIGC~\cite{zhou2024migc}, Instance~\cite{wang2024instancediffusion},
and RPG~\cite{yang2024mastering}.

\mypara{Datasets and Metrics.}
The T2I-CompBench \cite{huang2023ticompbench} is used to evaluate the MIS capability in terms of attribute binding, instance placement, and complex relationships using several criteria, including B-VQA ~\cite{pmlr-v162-li22n}, UniDet~\cite{Zhou_2022_CVPR}, and CLIP-Score~\cite{hessel2021clipscore}. As our method requires preliminary sketch information in advance, we utilize GPT-4V~\cite{openai2024gpt4technicalreport} to generate sketch images that automatically segment each instance and output the associated sketch mask (Details in \textbf{Appendix.\ref{app-section:5}}).
To further assess our method’s ability to handle extreme complexity, we customized 20 complex scenes with 6-8 instances and varying attributes to evaluate overall image quality (Details in \textbf{Appendix.\ref{app-section:6}}).
After the multimedia subjective testing~\cite{bt2002methodology}, we conducted a human assessment using a 1-5 rating scale for image quality, placement, and prompt-image consistency. 
Additionally, we used a pre-trained human preference model ImageReward~\cite{xu2023imagereward} to assess the aesthetic quality of images.

\subsection{Main Results}
\label{subsec:main}

\mypara{Quantitative Evaluation.}
We conduct quantitative evaluations on T2I-CompBench with upgraded sketches (\textbf{Appendix.\ref{app-section:5}}) and image quality assessment on customized complex cases (\textbf{Appendix.\ref{app-section:6}}) with study scores (1-5) and ImageReward~\cite{xu2023imagereward}, shown in Table~\ref{tab:main-results}. The results on T2I-CompBench indicate that our tuning strategy extends the base models with superior performance across all three scenarios, demonstrating exceptional generative capability in both layout and attribute synthesis and consistency with the findings in Section~\ref{subsec:analysis} regarding the unified attention mechanism. AST module refines cross-focus and self-focus attention for better prompt responsiveness, while HSLW tuning ensures each instance aligns with its corresponding attribute. 
The user study results in Table~\ref{tab:main-results} highlight improvements in generating human-preferred images for multi-instance scenarios over image quality, placement, and prompt-image consistency. 
Notably, ImageReward scores for aesthetic quality of images provide another insight that training-based/free methods typically show performance degradation relative to their baselines due to the distribution transferring, but our method maintains near-baseline performance with only slight decreases compared to SD v3.5 and FLUX.
These quantitative results collectively demonstrate our method's effectiveness in enhancing DiT-based models capabilities for accurate instance placement and attribute representation in MIS. 

\begin{wrapfigure}{r}{0.50\linewidth}
    \centering
    \includegraphics[width=\linewidth]{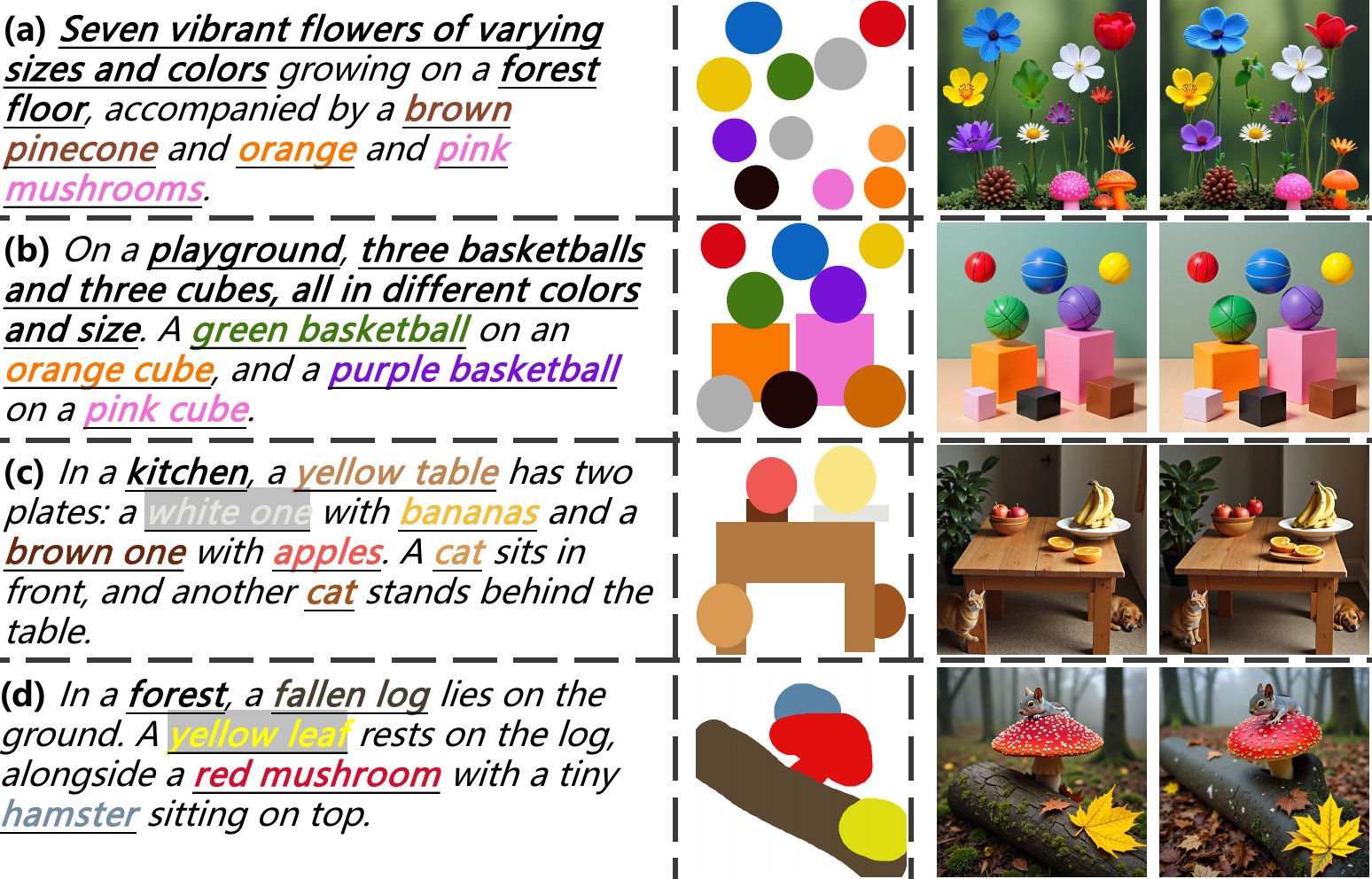}
    \caption{Visualization of handling small/repeated instances (``flower", ``mushroom", ``basketball") and occluded instances (``apples and bananas in plates", ``cat in front/behind table", ``leaf on log", ``hamster on mushroom").}
    \label{fig:flux-complex}
\end{wrapfigure}

\newpage
\mypara{Qualitative Evaluation.}
In the MIS comparison, we employ three complex prompts, each specifying over eight distinct instances with unique attributes (More examples are in \textbf{Appendix.\ref{app-section:6}}). Figure~\ref{fig:main} demonstrates that our method achieves superior precision in rendering multiple instances with designated attributes, maintaining sketch constraints and cohesive background-foreground integration. For instance, ``an orange church spire" in (a) and ``blue window" in (b) match the layout and specified color attributes precisely; similarly, ``blue cloud'' and ``green beach chair" in (c) conform precisely to the textual descriptions and layout constraints. 
Furthermore, we designed additional test cases featuring multiple repeated small instances and occluded instances to validate our method's capability in handling extreme complexity, as shown in Figure~\ref{fig:flux-complex}. The results demonstrate our method's effectiveness in: (1) numerous instances with varying sizes and colors (e.g., multiple flowers and differently sized/colored mushrooms in (a)); (2) complex spatial relationships (e.g., five basketballs with varying attributes in (b)); (3) intricate occlusion scenarios (e.g., cats before/after a table in (c), and a hamster on top of a mushroom in (d)).
These findings collectively demonstrate our method's robust capability for accurate instance placement and attribute representation under complex multi-instance sketch constraints.

\subsection{Ablation Study and Analysis}
\label{subsec:ab}

\mypara{Module Comparison.}

\begin{wrapfigure}{r}{0.50\linewidth}
    \centering
    \vspace{-40pt}
    \includegraphics[width=\linewidth]{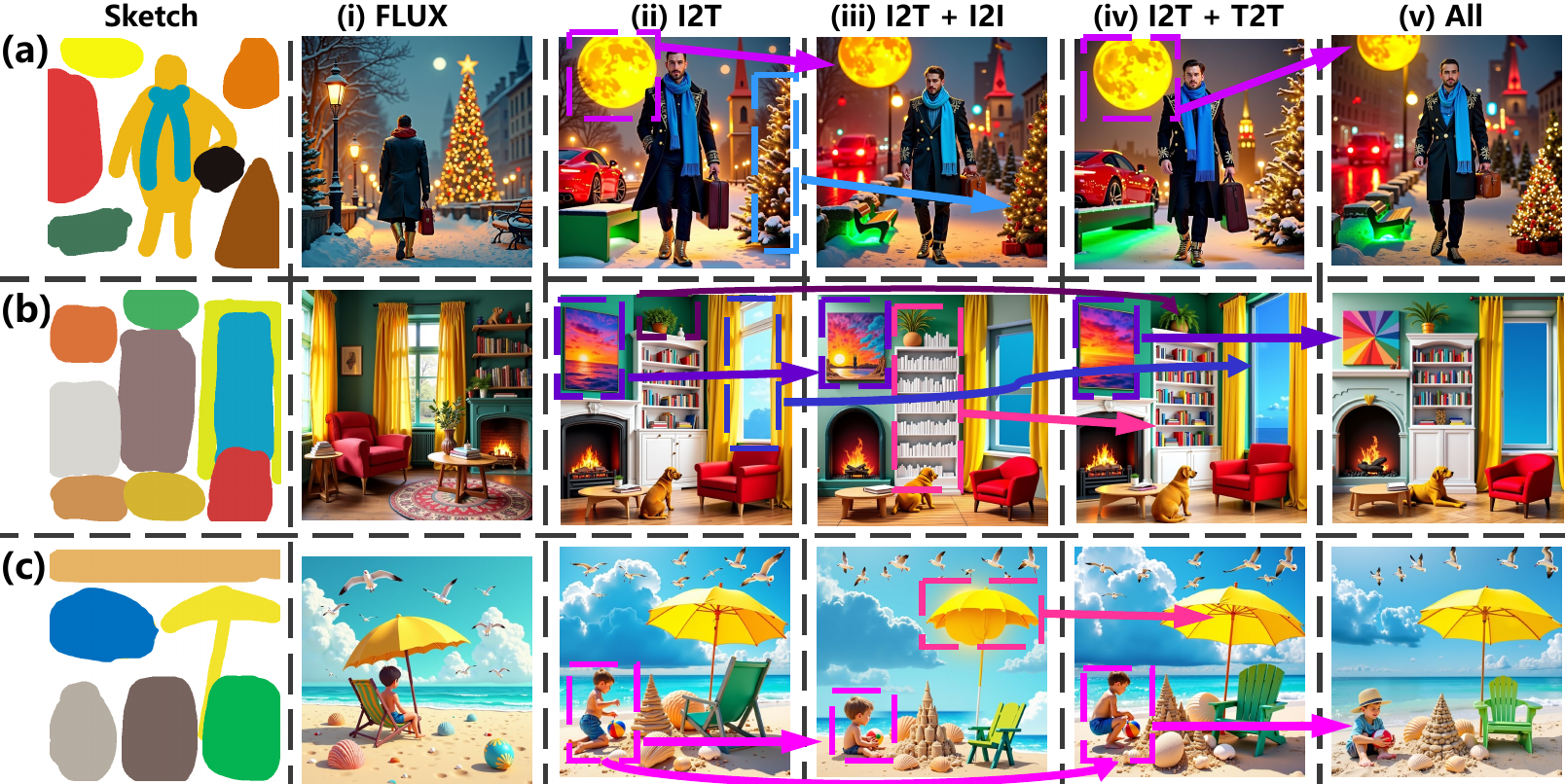}
    \caption{Ablation study on attention modules (Prompts in Figure~\ref{fig:main}). The I2T module mainly enhances layout-prompt alignment, T2T improves token interactions within segments, and I2I further refines instance self-alignment. Arrow direction indicates better. More study on hyper-parameters are in \textbf{Appendix.\ref{app-section:2}}.}
    \label{fig:ablation-one}
\end{wrapfigure}
Figure \ref{fig:ablation-one} shows visualization results, with attribute and position inaccuracies highlighted in colored boxes. The data reveals three main findings:
(1) The single I2T tuning strategy can generate all instances but lacks precise positioning, as seen in examples like the ``Christmas tree" in (a), and ``painting" in (b).
(2) Omitting the T2T module weakens token interactions at both segment and image granularity, affecting attention. For instance, without T2T, the ``boy" and ``ball" relationship in ``a boy playing with a beach ball" is disrupted in (c).
(3) Excluding the I2I region slightly degrades alignment with the sketch, as shown in (ii) and (iv), where nearly half of the instances deviate from their sketches. 
Besides the visualizations, we also conduct the quantitative evaluation on different variants presented in the last three rows of Table~\ref{tab:main-results}. While I2T alone improves all scenarios, T2T enhances token interactions, and I2I refines sketch-to-image alignment, both boosting human preference scores. These observations align with DiT-based analysis in Section~\ref{subsec:analysis}, emphasizing the essential roles of T2T, I2T, and I2I in prompt alignment.

\mypara{Analysis.} 

\begin{wrapfigure}{r}{0.50\linewidth}
    \centering
    \vspace{-10pt}
    \includegraphics[width=\linewidth]{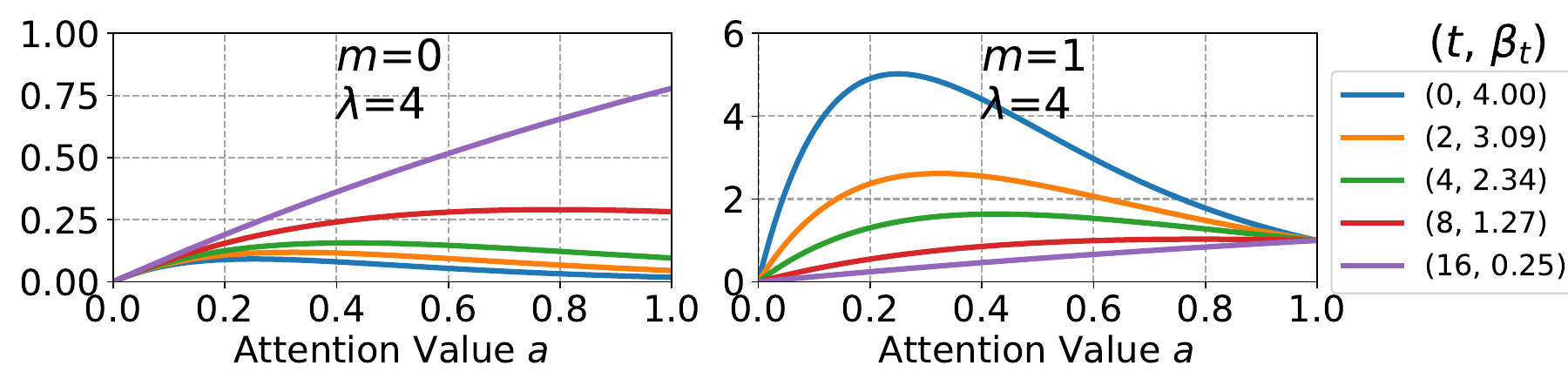}
    \caption{The scaling trend based on $
    a\cdot\exp \left( \beta_t
    \cdot(m-a) \right)$ of EQ.~(\ref{eq:atten}). $m=0$ means attenuating attention scores. $m=1$ means amplifying the attention score.}
    \label{fig:atten-tuning}
\end{wrapfigure}
We also examine the scaling trend based on $a\cdot\exp \left( \beta_t\cdot(m-a) \right)$ of Eq.~(\ref{eq:atten}) with $\lambda = 4.0$ and $T = 32$, considering two scenarios: $m = 0$ and $m = 1$, where $a$ represents the attention score $[0, 1]$. The results are shown in Figure~\ref{fig:atten-tuning}. For $m = 0$, all attention scores are suppressed, while for $m = 1$, the central point (\~{}0.2), aligns with the original range $[0, 0.3]$, refining the attention distribution. 
This phenomenon indicates that the unified scaling module suppresses less important regions and enhances key areas.

\section{Conclusion}
In this work, we investigated the attention mechanisms in the DiT architecture, analyzing various attention modules, and exploring the hierarchical attention responses across different tokens with step-layer-wise token exchange experiments. 
We proposed a training-free approach with hierarchical and step-layer-wise attention specialty tuning, enabling precise MIS using the DiT-based models. Both quantitative and qualitative evaluations on upgraded sketch-based T2I-CompBench and customized complex cases showed that our approach achieved accurate instance placement and attribute representation in detailed multi-instance layouts. Moreover, we established an analysis-design-experiment framework that can serve as a potential paradigm for future improvements in MIS. 

\bibliographystyle{unsrt}  
\bibliography{main}  

\newpage

\section*{\MakeUppercase{Supplementary Materials}}
\renewcommand{\thesection}{\arabic{section}}
\setcounter{section}{0}

\section{Ethical and Social Impacts}
\label{app-section:1}
\begin{figure}
        \centering
        \includegraphics[width=0.65\linewidth]{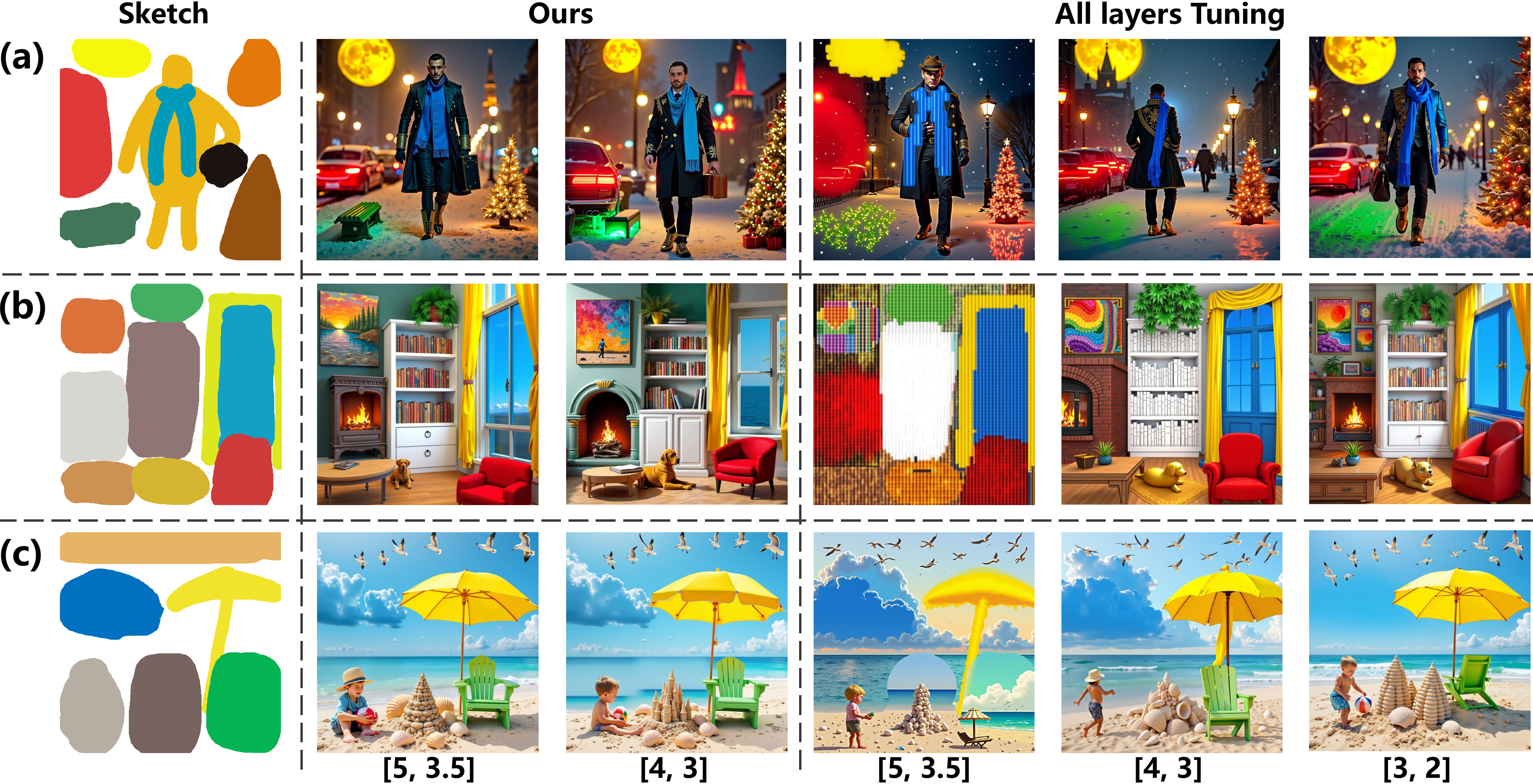}
        \caption{Ablation on step-layer-wise tuning. Full-layer tuning: adding tuning across all layers with varying $[\lambda_{cross}, \lambda_{self}]$ degrades image quality, leading to instance repetition, over-saturation, and distortion due to disrupted attention responses.}
        \label{fig:ablation-two}
\end{figure}
This study on MIS in DiT, which incorporates AST and HSLW modules and builds on the FLUX and SD v3.5 base models from the HuggingFace diffusers library~\cite{blackforestlabs_flux_2024}, assesses performance on the widely used T2I-CompBench datasets~\cite{kaiyi2024t2ibench} and raises several ethical and societal implications. Primary among these is the privacy preservation challenge in synthetic media generation, as the model's capability to synthesize multi-instance compositions from textual descriptions potentially enables the unauthorized reproduction of individual likenesses, necessitating robust privacy-preserving mechanisms. 
In addition, the underlying FLUX and SD v3.5 architectures exhibit inherent representational biases that can manifest themselves in stereotypical image generation, requiring continuous bias monitoring and mitigation strategies to ensure equitable representation. 
While our methodology advances diverse multi-instance figure generation, it also poses risks regarding the creation of synthetic media for malicious purposes. 
This research demonstrates that advancing DiT capabilities must be accompanied by rigorous ethical considerations, particularly in privacy preservation, bias mitigation, and the establishment of safeguards against misuse in multi-instance figure synthesis.

\section{Hyper-Parameters $\lambda$ and Layer Comparison}
\label{app-section:2}
\begin{figure}[th]
    \centering
    \includegraphics[width=0.8\linewidth]{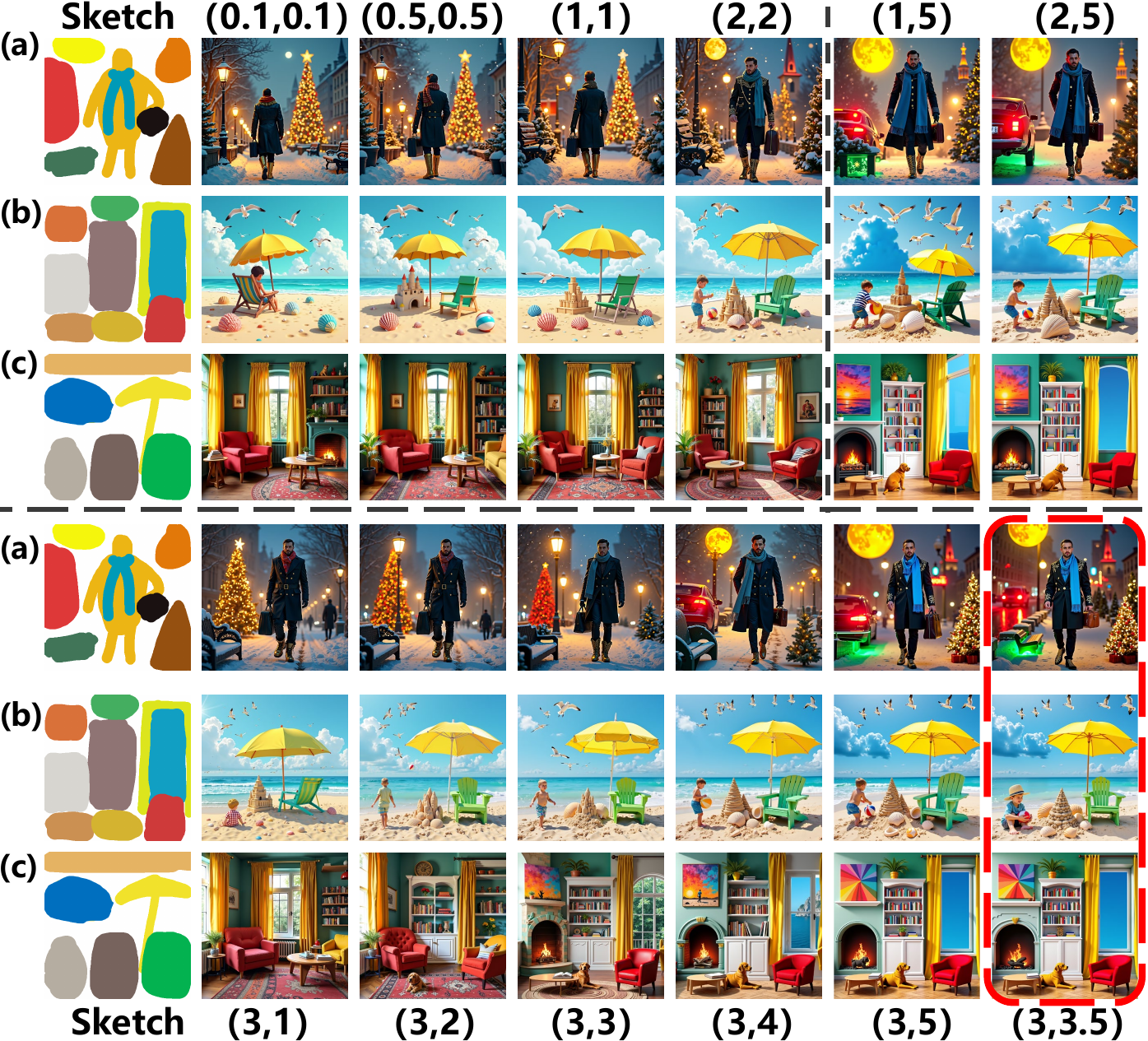}
    \caption{The visualization comparison with different hyper-parameters $(\lambda_{self},\lambda_{cross})$  on complex MIS. The results of our choices  present the best performance on the the instance placement and attribute representation.}
    \label{fig:hps}
\end{figure}
In this section, we systematically investigate the impact of two critical design elements: the hyper-parameters 
$(\lambda_{self},\lambda_{cross})$ in the AST module and our proposed HSWL optimization strategy. First, we conduct thorough experiments to determine optimal hyper-parameter $(\lambda_{self},\lambda_{cross})$ in the AST module, formalized in Eq. (2) of the main text, and configurations for MIS. Through empirical analysis of both symmetrical parameter pairs [(0.1, 0.1), (0.5, 0.5), (1, 1), (2, 2)] and asymmetric configurations [(1, 5), (2, 5), (3, 1)-(3, 5)], we demonstrate that the optimal parameter combination (3.5, 5) achieves superior synthesis quality. As shown in Figure~\ref{fig:hps}, this configuration maintains precise instance localization while ensuring visual coherence.

To validate the necessity of our HSLW module, we conduct a comparative full-layer tuning experiment with results shown in Figure \ref{fig:ablation-two}. The full-layer tuning behaves as an over-manipulation of the attention map, causing significant distortion in both foreground and background. With different parameters, the visualization results reveal several issues, including instance repetition (``painting" in (b) and ``sandcastle" in (c)), missing instances (``green bench" in (a) and ``blue cloud" in (c)), attribute leakage (``green plant" in (b)), and inaccurate placement across all three cases. These suboptimal results confirm the effectiveness of our layer-tuning strategy, which tunes the attention map via specific masks across different diffusion phases, enhancing coherence and accuracy.

\section{Detailed Formulation of G} 
\label{app-section:3}
To compute the similarity matrix $G$, we first transform each binary sketch mask $S_i \in \mathbb{R}^{h \times w}$ into a flattened vector representation $\bar{S}_i \in \mathbb{R}^{hw \times 1}$. Given a batch of sketch masks, we construct an intermediate correlation matrix $L\in \mathbb{R}^{hw \times hw}$ through batch-wise outer products:
\begin{equation}
L = \sum_{i=1}^{batch}\bar{S}_i \bar{S}_i^T 
\end{equation}

The sensitivity matrix $G\in \mathbb{R}^{hw \times 1}$ is then computed by normalizing and scaling $L$:
\begin{equation}
G = 1 - \gamma \frac{\sum_{i=1}^{hw} L_i}{hw}
\end{equation}
where $\gamma$ is a modality-dependent scaling factor set to 4.0 for text queries and 1.0 for image queries. This formulation guarantees that $G$ captures the pairwise relationships between spatial locations while taking into account the characteristics specific to the modality.

\begin{figure*}[thb]
    \centering
    \includegraphics[width=0.90\textwidth]{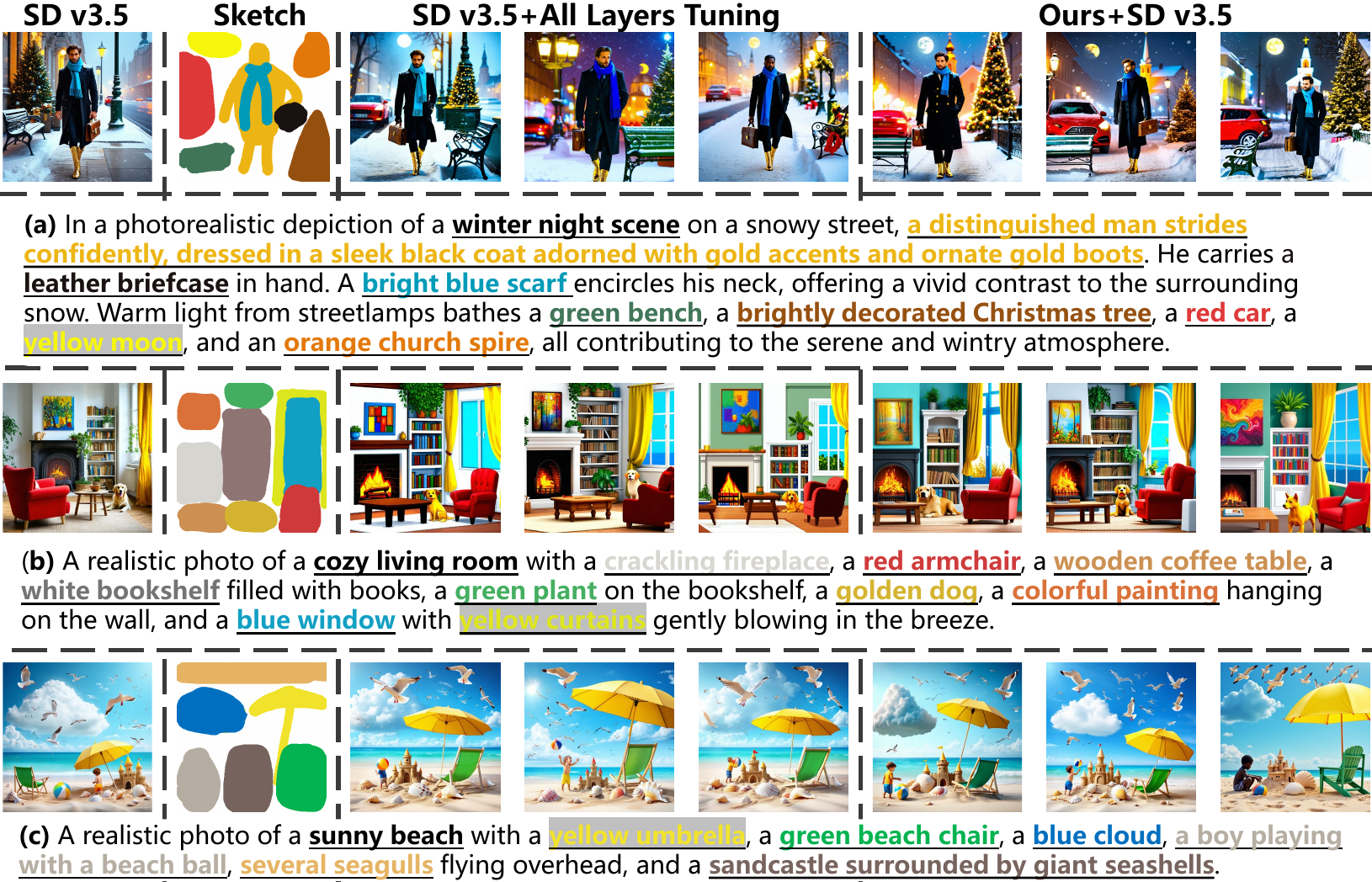}
    \caption{Visualizations for transferring our approach to the SD v3.5 model~\cite{esser2024scaling}. The corresponding prompts are also shown in Figure~5 in the main text and more are shown by Figures~\ref{fig:app-main-figures1}-\ref{fig:app-main-figures3} in Appendix.5. We present two implementation variants: The first variant integrates the AST module with a full-layer tuning strategy, i.e., SD v3.5+All Layers Tuning, whereas the second variant combines the AST module with a HSLW tuning strategy while retaining the default configuration, i.e., Ours+SD v3.5. When implemented on SD v3.5, our framework achieves precise multi-instance alignment between textual prompts and spatial layouts, as shown by the visualization results. The underlined prompts within the prompts are the sub-prompts for instances, with colors matching those in the sketch.}
    \label{fig:sd3}
\end{figure*}

\section{Module Comparison on SD v3.5 Model} 
\label{app-section:4}
To validate the general applicability of our HSLW module beyond the FLUX model, we apply our approach to another flow-matching DiT-based model, Stable Diffusion 3.5 (SD v3.5)~\cite{esser2024scaling}. Since SD v3.5 may exhibit step- and layer-specific behavior across its four attention areas, we implement the same HSLW attention tuning strategy during the first half of the steps for effective validation. SD v3.5 concatenates text embeddings from CLIP and T5 tokenizers, resulting in two corresponding T2I and T2T masks per tokenizer, applied as outlined in Figure 4 of the main text. Using the same prompts and sketches as in Figure 5 of the main text, we adopt the same hyperparameters except for $(\lambda_{self}, \lambda_{cross})$, adjusted to avoid over-manipulation of the attention maps. As demonstrated in Figure~\ref{fig:sd3}, HSLW module significantly improves sketch-image alignment compared to standalone AST modules, while preserving prompt consistency across multi-instance scenarios. This confirms the model-agnostic compatibility of AST+HSLW modules with flow-matching DiT-based models, achieving performance comparable to FLUX-dev implementations. Future research will investigate independent HSLW applications to both CLIP- and T5-derived attention maps. Despite these limitations, the results indicate effective generalization and potential for advancing future flow-matching DiT-based  models.


\section{Upgraded Sketch of T2I-CompBench}
\label{app-section:5}
Our method requires preliminary sketch information, which we obtain using GPT-4V~\cite{openai2024gpt4technicalreport}, a most-recent language-vision model. GPT-4V creates sketch images by automatically segmenting each instance and outputting the corresponding sketch mask based on a textual description of the desired scene. To maintain proper spatial layouts and relationships, all generated sketch images are manually refined. The sketch source, including the GPT-4V generated sketches and their manually adjusted counterparts, will be accessible upon the release of the main code.

Meanwhile, using T2I-CompBench evaluation metrics, i.e., B-VQA~\cite{pmlr-v162-li22n}, UniDet~\cite{Zhou_2022_CVPR}, and CLIP-Score~\cite{hessel2021clipscore}, all yield scores of 0 (range: 0 to 1, with 1 being best), resulting our method showing a slight increase in Table 1 of the main paper. To illustrate this failure, we randomly select six cases (three spatial, three color) displayed in Figure~\ref{fig:t2i-six}. Upon manual inspection of the images generated from T2I-CompBench prompts, most comply with the attribute representation and layout constraints.  The problem arises from a distribution mismatch: these evaluation tools were trained on data significantly different from our generated images, resulting in lower T2I-CompBench scores~\cite{NEURIPS2023_f4d4a021}. This failure of pre-trained models to effectively process generated images is well-documented~\cite{Epstein_2023_ICCV,Wang_2023_ICCV}. Therefore, we also designed more customized complex cases to evaluate image quality in the next section.



\section{Customized Complex Cases}
\label{app-section:6}
\begin{figure}
    \centering
    \includegraphics[width=0.65\linewidth]{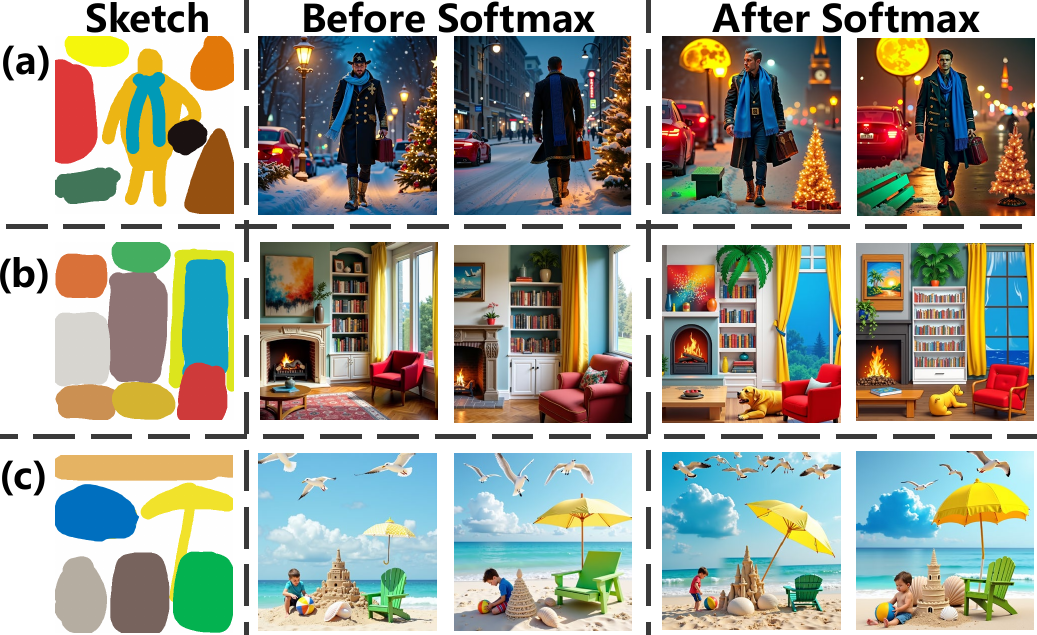}
    \caption{Comparison with attention modulation before/after softmax. Applying the unified scaling module to the attention map after softmax effectively enhances alignment between prompts and sketch masks.}
    \label{fig:denseflux}
\end{figure}
To evaluate the overall image quality, we customized 20 complex scenes containing 6-8 instances and invited the participants to evaluate our prompt-generated image pairs under the guidelines of the multimedia subjective
testing~\cite{bt2002methodology} from the perspective of prompt-image alignment, sketch/layout-to-image coherence, aesthetic preference. Each case specifies more than 6 distinct instances with unique attributes. As depicted in Figures~\ref{fig:app-main-figures1}-\ref{fig:app-main-figures3}, our method produces multiple instances with their designated attributes more accurately, cohesively integrating background and foreground elements while adhering to the sketch constraints. The misalignment between the textual description and the prior sketch information in the other models leads to these limitations.

We implement a pipeline for systematic assessment: (1) Fair Generation: For each of 20 complex scenes, each method generates 20 images with random seeds.  (2) Random Sampling: From each scene-method pair's 20 variants, we randomly select one evaluation instance. (3) Blinded Assessment: All selected samples are aggregated and presented in randomized order, preventing positional bias during assessment. For the evaluation, we require the participants to consider the explicit spatial relationship between segments regarding the layout. To be specific, the criteria for overall assessment could be summarized as: 
\begin{itemize}
    \item \textbf{5:} Alignment with both layout and textual prompts. All elements are correctly placed with high quality (sharp details, proper proportions, no artifacts).
    \item \textbf{4:} Minor discrepancies in 1 element missing or misplacing, or slight quality degradation (blurring/mild distortion) in one component.
    \item \textbf{3:} More errors in 2 elements omissions or misplacements, or moderate quality issues (visible distortion/partial incoherence).
    \item \textbf{2:} More than 3 placement inaccuracies, or errors affecting more than 2 elements with quality flaws (severe distortions/unnatural blending).
    \item \textbf{1:} More than 4 errors elements missing or severely misplaced with low-quality rendering (fragmented objects/unrecognizable shapes).
\end{itemize}

\section{AST Module Before and After Softmax} 
\label{app-section:7}
To validate our AST module, which modulates the attention map after softmax, we qualitatively compare the visualization results with the attention modulation method in Dense Diffusion as shown in Eq. (2) \cite{kim2023dense}. The visualization results are provided in Figure~\ref{fig:denseflux}. These results indicate that attention modulation before softmax presents several issues, including incorrect instances (``scarf" in (a) and ``yellow umbrella" in (c)), missing instances (``green bench" in (a) and ``golden dog" in (b)), inaccurate attributes (``blue cloud" in (c)), and misplacement in all three cases. These suboptimal outcomes validate our AST strategy, which modulates the attention map after softmax, enhancing the coherence and accuracy of MIS.

\section{More Attention Analysis} 
\label{app-section:8}

\subsection{Analysis of T2I Attention Map} 
\begin{wrapfigure}{r}{0.50\linewidth}
    \centering
    \includegraphics[width=0.85\linewidth]{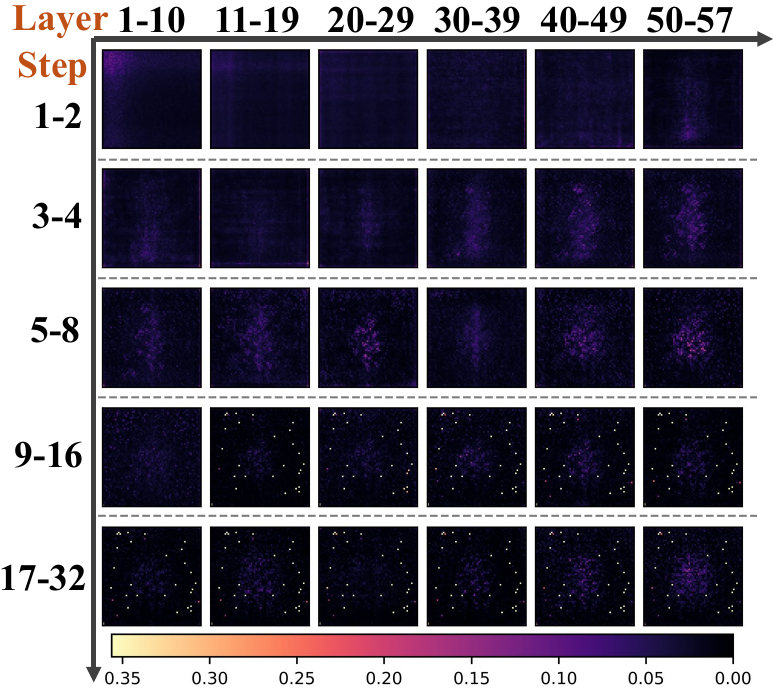}
    \caption{The T2I average attention maps omitted in Section 3.2.  Compared to other attention maps, T2I maps exhibit lower scores, indicating a weak impact, and consequently, they are not modulated by our AST module.}
    \label{fig:attn-T2I}
\end{wrapfigure}

This part presents the T2I maps from Section 3.2, which were omitted from the main text due to space constraints. These maps are depicted in Figure~\ref{fig:attn-T2I}. In comparison to other attention modules, T2I maps exhibit significantly lower attention scores and contain limited semantic information, resulting in a weak influence on the overall attention map. In conclusion, while T2I maps could slightly improve text responsiveness, they significantly increase computational burden and are therefore not incorporated into our method.

\subsection{More Attention Analysis of FLUX-dev}
We observed similar patterns in attention maps across multiple instances, but we only include one prompt example in the main paper due to space constraints. Here, we provide additional map analysis in the full paper, focusing on the prompt ``Yellow orb in cave" to systematically assess four attention regions within the FLUX framework. The visualization maps are categorized by types as shown in Figure~\ref{fig:attn-me}: self-attention maps (T2T and I2I), and cross-attention-based I2T and T2I maps, supporting the consistent findings presented in Figure 2 of the main paper. Additionally, we performed a t-SNE analysis of attention maps across different text prompts, as demonstrated below.

\subsection{t-SNE Analysis of Attention}
\begin{figure*}[t]
    \centering
    \includegraphics[width=0.95\textwidth]{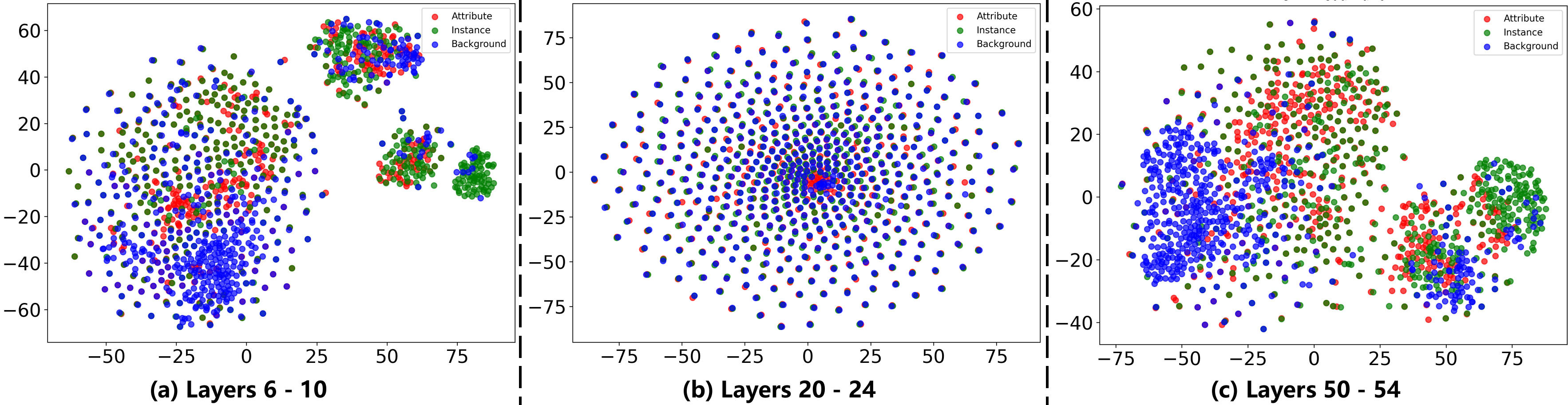}
    \caption{Systematic analysis of Flux model's attention mechanisms using 512 GPT-4-generated structured prompts (``[color][instance] in [background]"), e.g.,  ``Green pyramid in the ocean", ``Yellow cone in the mountain", and ``Purple cylinder in the river". In the shallow layers (6-10), instances exhibit a more aggregated pattern, with attributes and backgrounds displaying a more dispersed distribution. This highlights the importance of instances in these layers. In the middle layers (20-24), all three classes—instances, attributes, and backgrounds—are more dispersed from each other, with no clear aggregation. This suggests a significant interplay between them. Notably, while the background tends to aggregate in other layers, it becomes more intertwined with other elements in the middle layers, confirming that background control has varied effects at this stage. In the deep layers (50-54), the separation and aggregation of backgrounds and instances become more pronounced, alongside the strong coupling between attributes and instances. This indicates that attributes play a crucial role in the later stages.}
    \label{fig:tsne}
\end{figure*}

\begin{figure*}[!htb]
    \centering
    \includegraphics[width=0.95\textwidth]{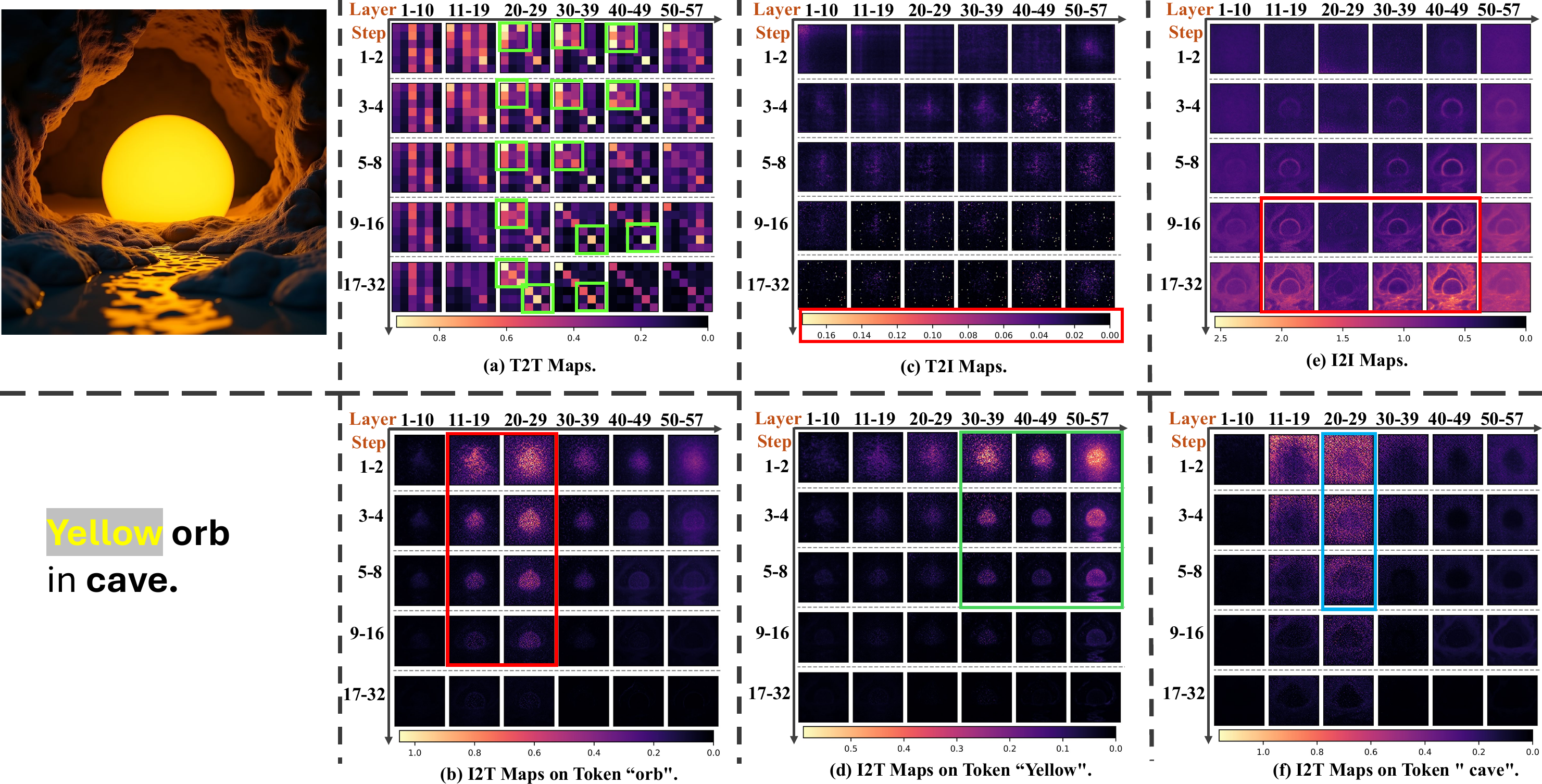}
    \caption{Attention map of the FLUX-dev model averages for the prompt ``Yellow orb in cave.", which strengthen our findings in Section 3.2. (a) T2T maps show strong intra-segment interactions within valid tokens of ``Yellow orb" (first 3 tokens) and ``cave" (last 2 tokens). (b) I2I maps (the maximum value of each point relative to all other points) reveal consistent interaction between ``Yellow orb" and ``cave" in the middle layers of four steps. (c) T2I maps show lower scores relative to others, indicating a weak impact. (d/e/f) I2T attention maps on individual tokens highlight that instance tokens dominate early layers, background tokens in the middle, and color tokens later, with most information integrated in the first half of steps.}
    \label{fig:attn-me}
\end{figure*}

To comprehensively analyze the attention mechanism in the FLUX model, we conducted experiments using diverse textual prompts and performed statistical analysis of attention responses across multiple samples. Specifically, we employed GPT-4 to generate 512 structured prompts in the format ``[color] [Instance] in [Background]" to ensure systematic coverage of variations.
For a detailed investigation of attention response patterns, we extracted statistical summaries by aggregating attention maps across three key layer ranges (6-10, 20-24, and 50-54) in the first 16 steps. These layer ranges were chosen based on preliminary analysis of attention distribution in Section 3.2. We then applied t-SNE dimensionality reduction to visualize and analyze the characteristic patterns of attention responses, enabling robust verification of our findings regarding the attention distribution across different components of prompts. The results of the t-SNE analysis on three chosen layers are displayed in Figure~\ref{fig:tsne}. In the shallow layers (6-10), instances exhibit a more aggregated pattern, while attributes and backgrounds display a more dispersed distribution. This highlights the significance of instances in these layers. In the middle layers (20-24), all three classes—instances, attributes, and backgrounds—are more dispersed from each other, with no clear aggregation. This suggests a considerable interplay among them. Interestingly, while the background tends to group in other layers, it becomes more intertwined with other elements in the middle layers, confirming that background control has varied effects at this stage. In the deep layers (50-54), the separation and aggregation of backgrounds and instances become more evident, alongside a strong coupling between attributes and instances. This indicates that attributes are crucial in the later stages.

\subsection{Attention Analysis of FLUX-schnell} 
\begin{figure*}[!htb]
    \centering
    \includegraphics[width=0.95\textwidth]{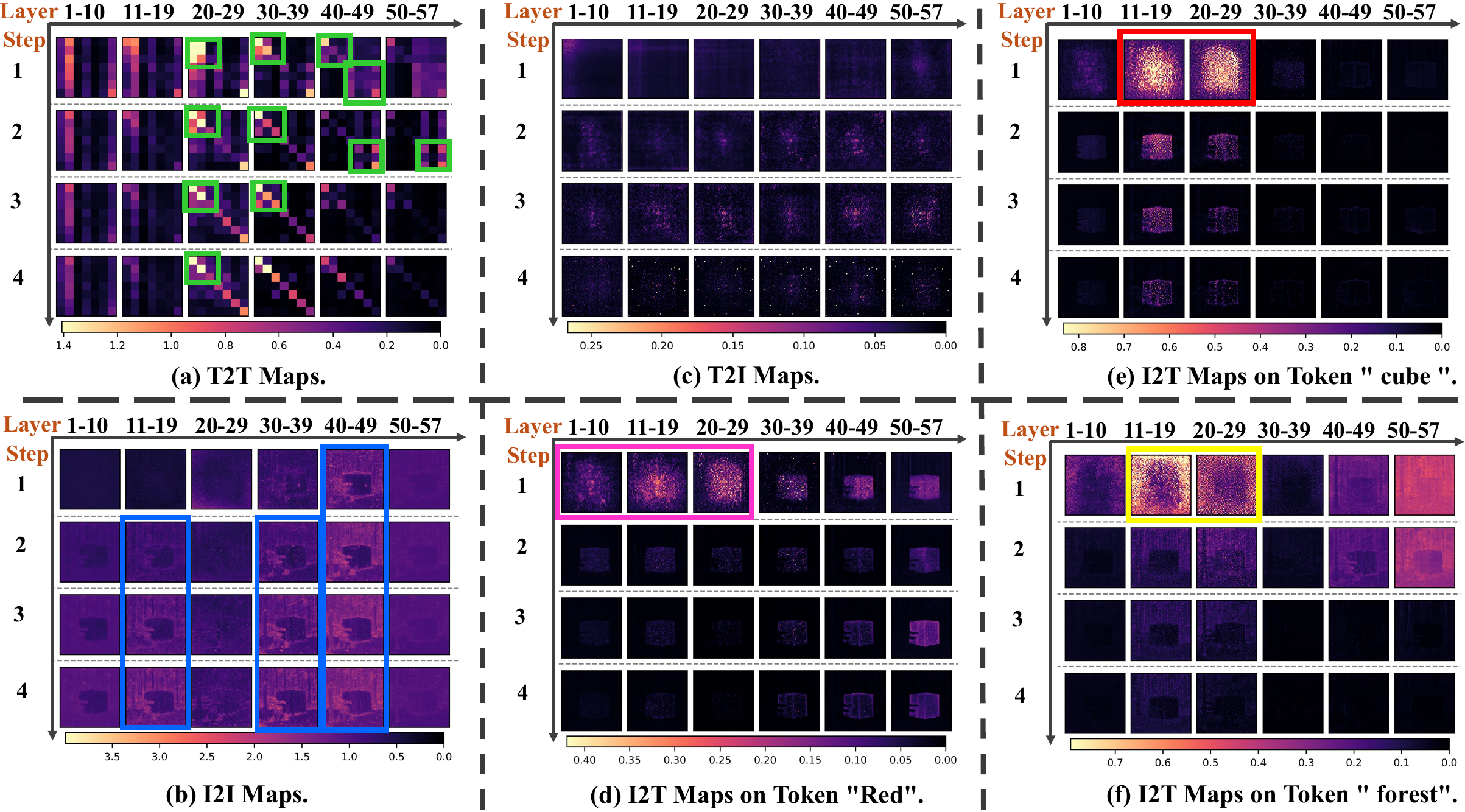}
    \caption{Attention map of the FLUX-schnell model averages for the prompt ``Red cube in a forest", which illustrates similar findings to the FLUX-dev model in Section 3.2. (a) T2T maps show strong intra-segment interactions within valid tokens of ``Red cube" (first 3 tokens) and ``in a forest" (last 4 tokens). (b) I2I maps (the maximum value of each point relative to all other points) reveal consistent interaction between ``Red cube" and ``a forest" in the middle layers of four steps. (c) T2I maps show lower scores relative to others, indicating a weak impact. (d/e/f) I2T attention maps on individual tokens highlight that instance, background, and color tokens all dominate in the left half layers of the first step.}
    \label{fig:attn-schnell}
\end{figure*}

Following the same analytical process in Section 3.2, we employ the prompt ``Red cube in a forest" to examine the functionality of four regions in the attention maps of the FLUX-schnell model. The visualization maps are categorized by types, as depicted in Figure~\ref{fig:attn-schnell}, confirming the observations made in the FLUX-dev model analysis.

\section{Details of Token Exchanging Experiments} 
\label{app-section:9}
\begin{figure*}[!htb]
    \centering
    \includegraphics[width=0.95\textwidth]{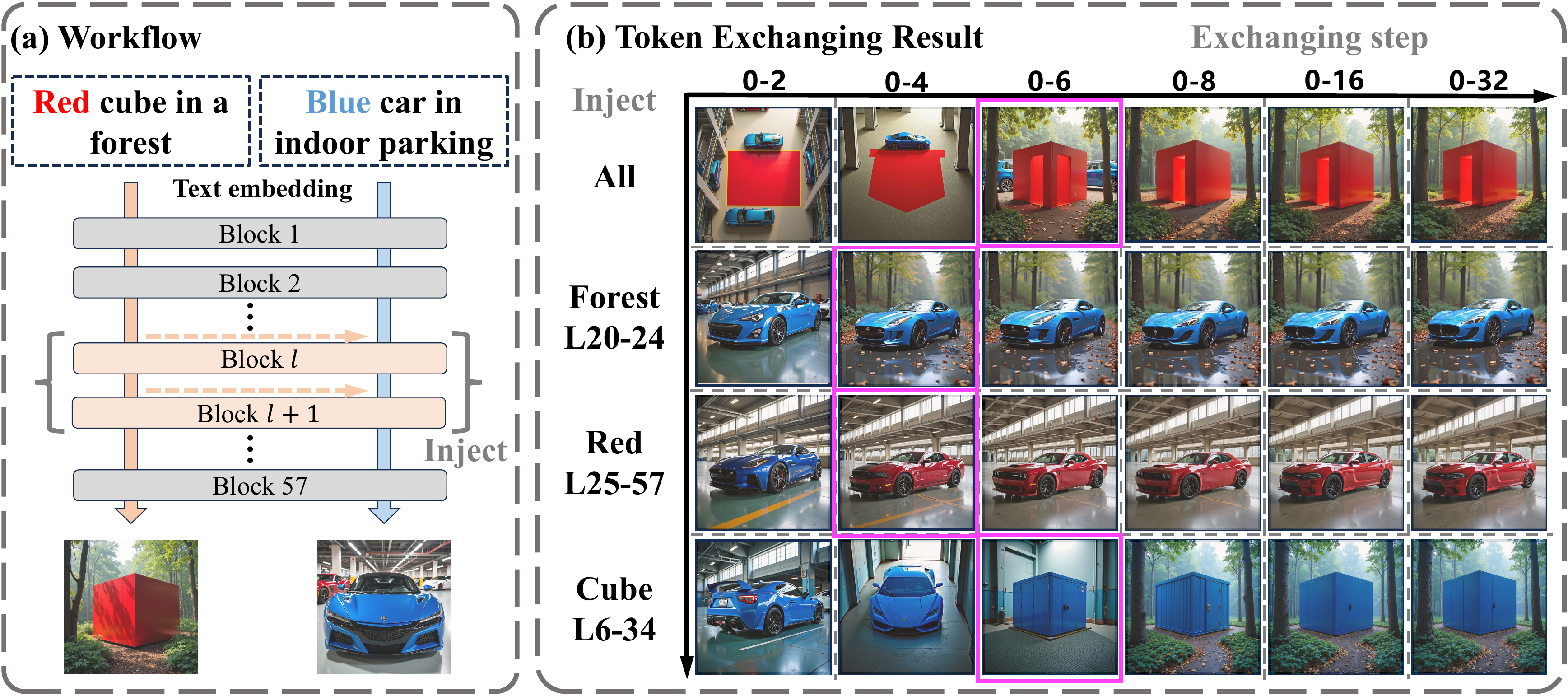}
    \caption{Token exchanging experiment between the prompts ``Red cube in a forest” and ``Blue car in indoor parking”
    (a) In a single batch, we exchange the token-specific elements of the text embedding between the left and right sides at specific layers and steps.
    (b) Results reveal that during the initial 8 steps, it is possible to selectively replace specific token components at certain layers. These controlled exchanges support the hierarchical attention response patterns observed in Figure 2 of the main text, providing further evidence for the model's ability to handle fine-grained manipulations.}
    \label{fig:token-exchange-three}
\end{figure*}

\begin{figure*}[!htb]
    \centering
    \includegraphics[width=0.95\textwidth]{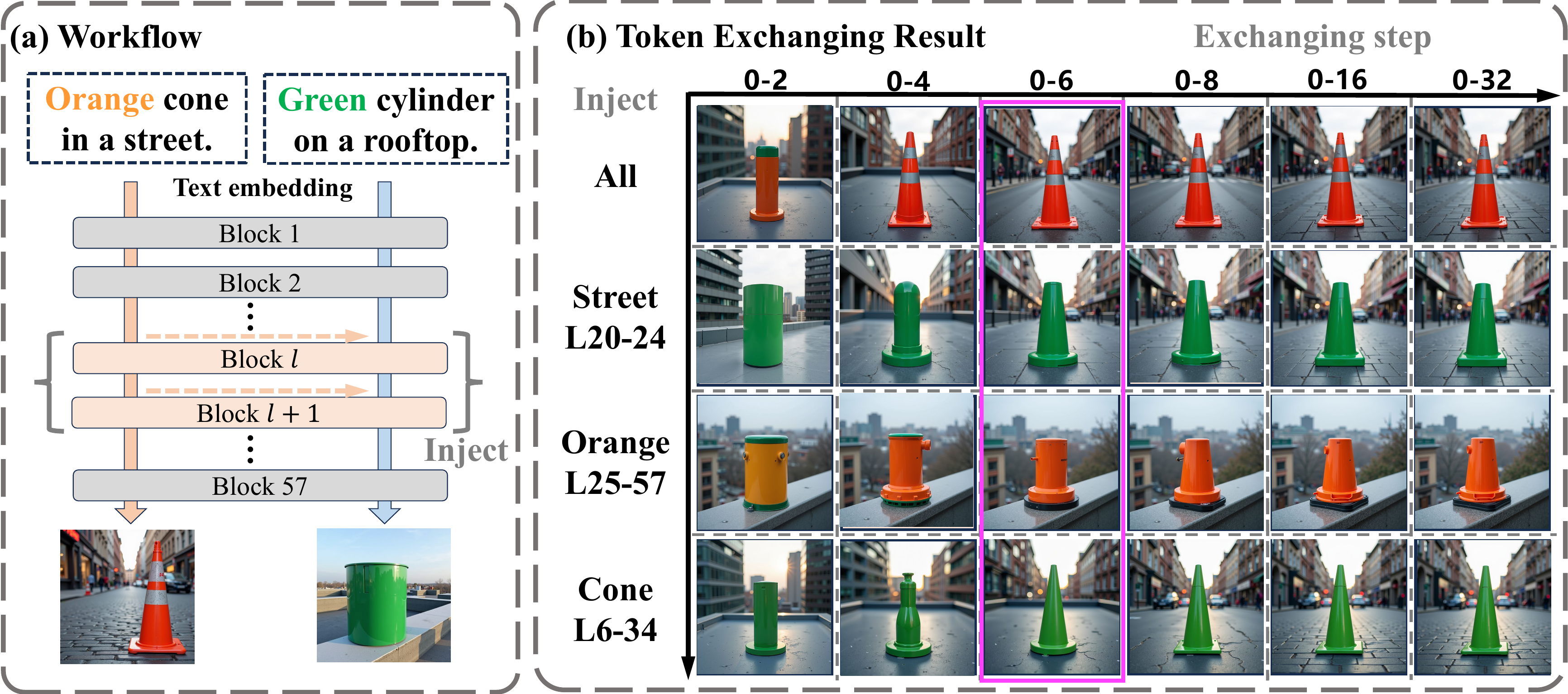}
    \caption{Token exchanging experiment between the prompts ``Orange cone in a street" and ``Green cylinder on a rooftop".
    (a) In a single batch, we exchange the token-specific elements of the text embedding between the left and right sides at specific layers and steps.
    (b) Results reveal that during the initial 8 steps, it is possible to selectively replace specific token components at certain layers. These controlled exchanges support the hierarchical attention response patterns observed in Figure 2 of the main text, providing further evidence for the model's ability to handle fine-grained manipulations.}
    \label{fig:token-exchange-one}
\end{figure*}

\begin{figure*}[!htb]
    \centering
    \includegraphics[width=0.95\textwidth]{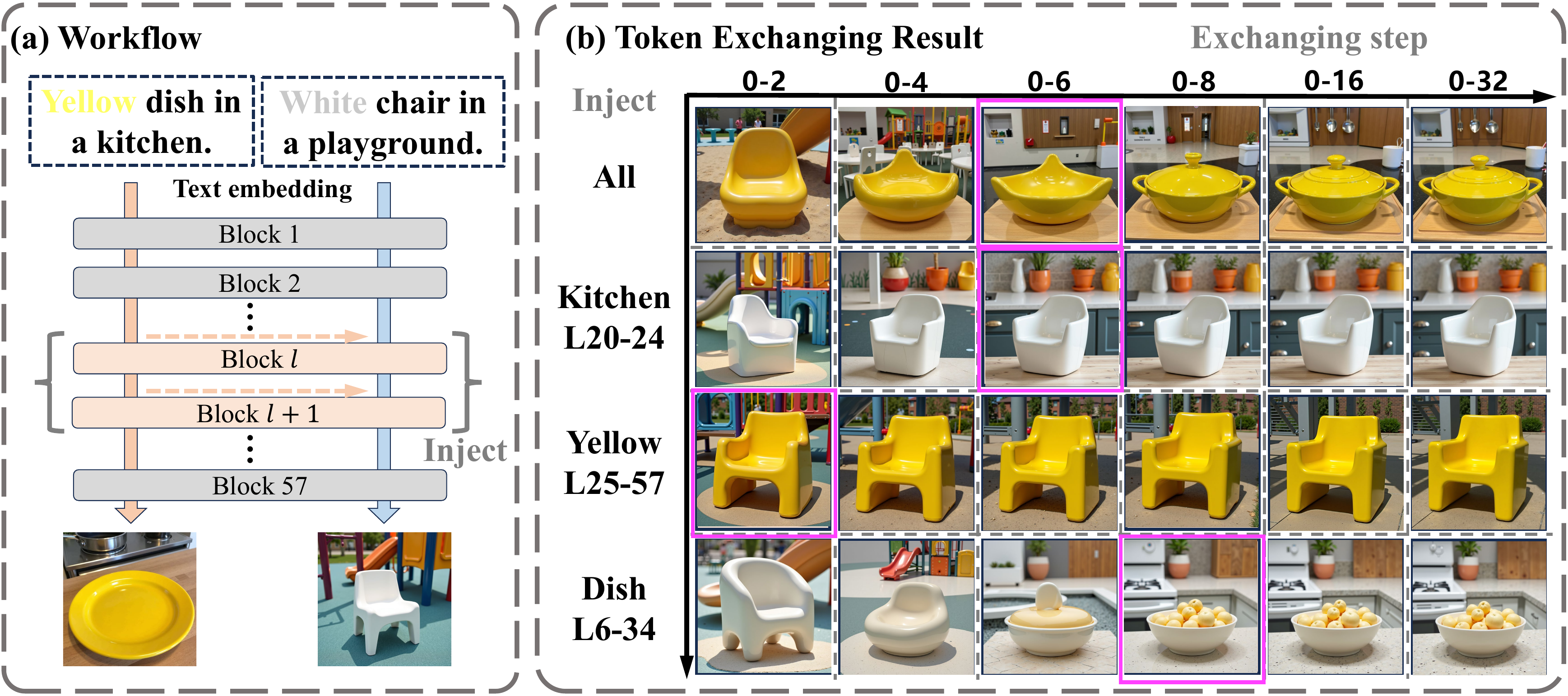}
    \caption{Token exchanging experiment between the prompts ``Yellow dish in a kitchen" and ``White chair in a playground".
    (a) Within the same batch, we swap the token-specific components of the left side's text embedding with their right-side counterparts at designated layers and steps.
    (b) Findings indicate that specific token components can be strategically substituted at designated layers within the first 8 steps. These targeted replacements corroborate the hierarchical attention responses depicted in Figure 2 of the main text, reinforcing the model's capacity for precise control.}
    \label{fig:token-exchange-two}
\end{figure*}

\begin{figure*}[b]
    \centering
    \includegraphics[width=1.0\textwidth]{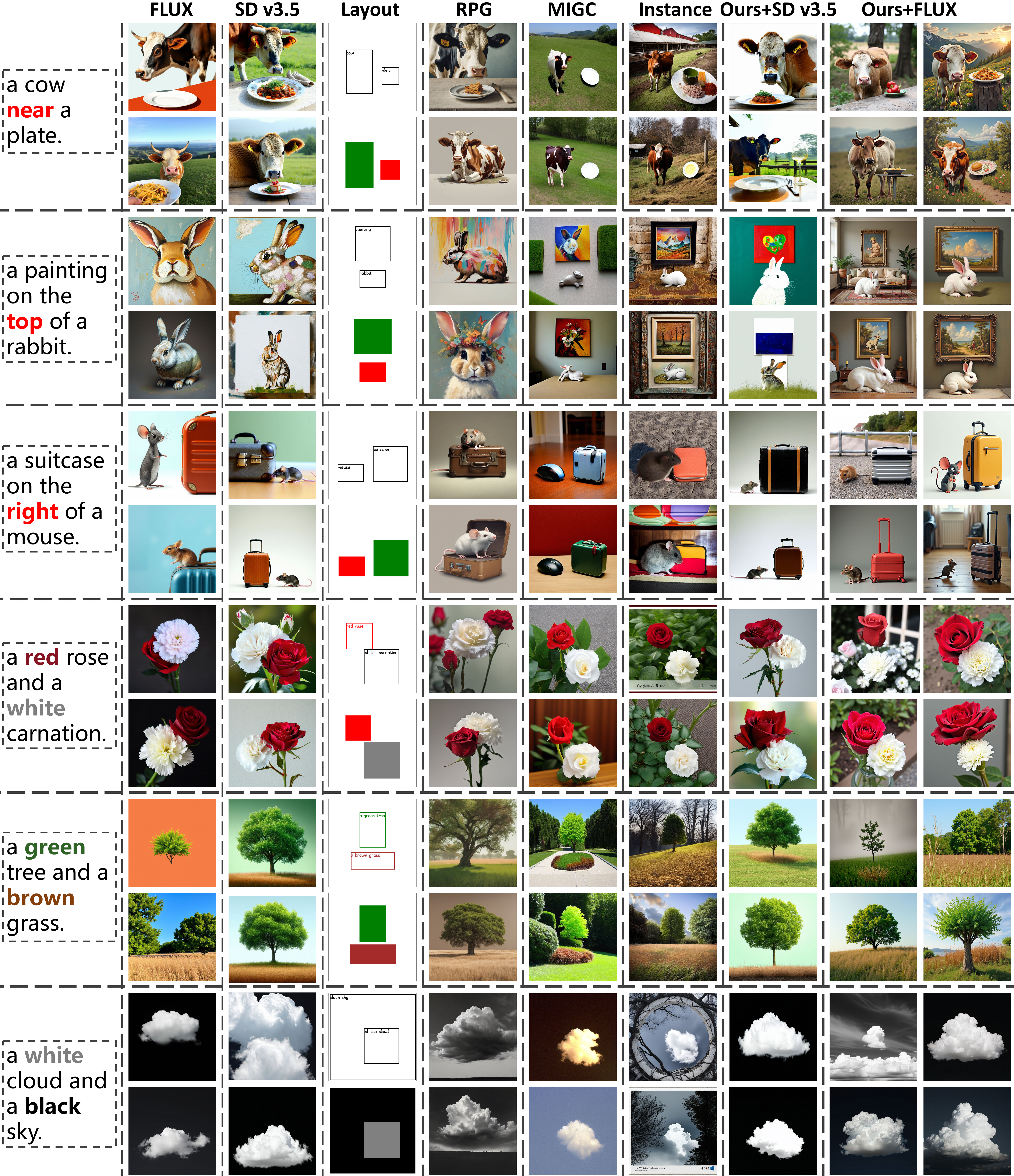}
    \caption{Comparison with other diffusion models on six T2I-CompBench. Although our method has achieved accurate instance placement, attribute representation, and enhanced diversity and realism in image quality, these generated images all received a zero score instead of the correct score. These failure cases (at least 10\% in each category) contribute to our relatively low performance in the T2I-CompBench, which also indicates that a more comprehensive and powerful benchmark is required for the current text-to-image diffusion models, whose data distribution greatly deviates from the training distribution of the T2I-CompBench evaluation tools used.}
    \label{fig:t2i-six}
\end{figure*}

\begin{figure*}[b]
    \centering
    \includegraphics[width=0.95\textwidth]{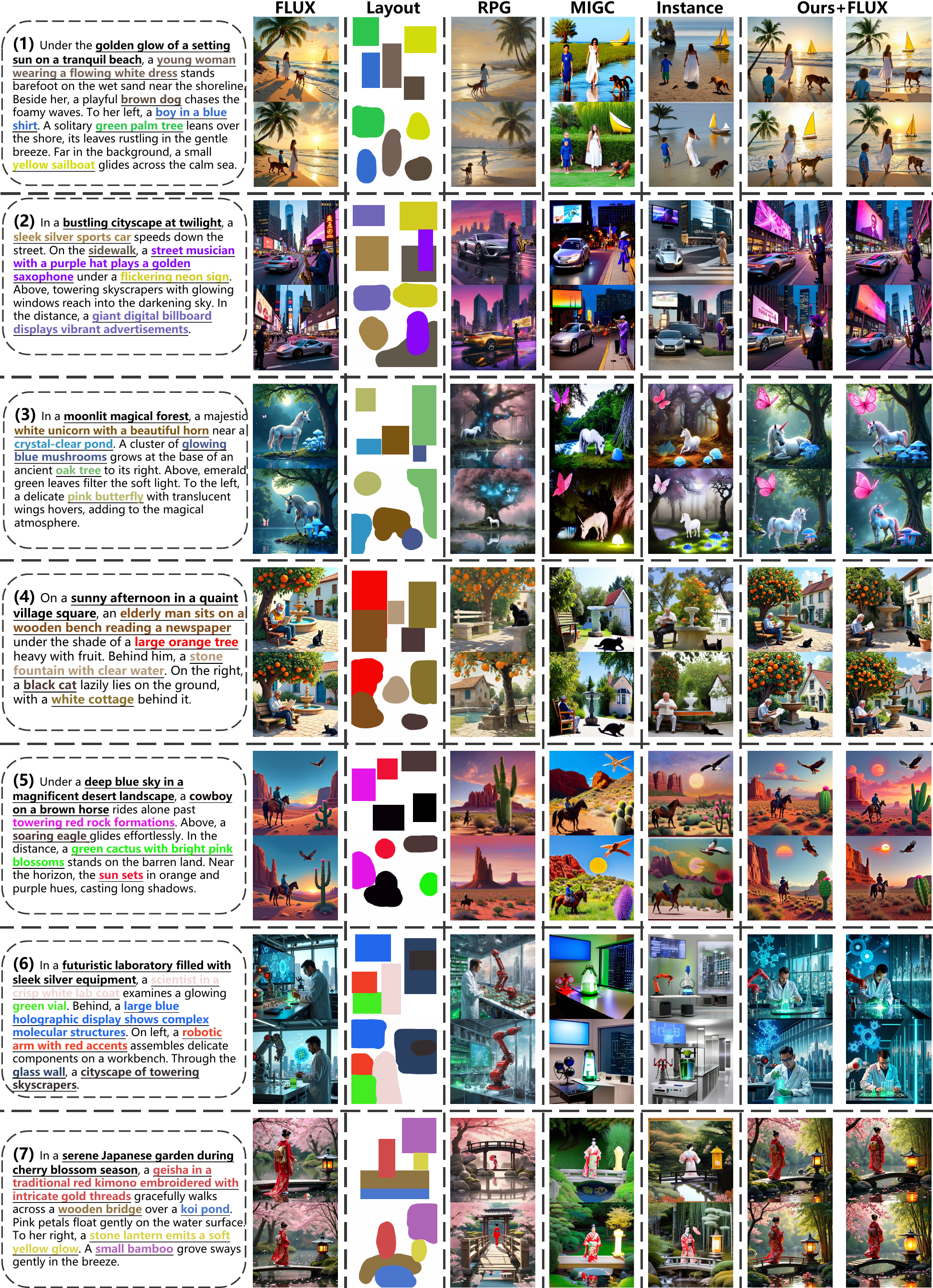}
    \caption{Comparison of 7 cases with other diffusion models, supplementing the main result (Figure 5 in the main text). Bounding box layouts are provided for MIGC~\cite{zhou2024migc} and InstanceDiffusion~\cite{wang2024instancediffusion}, while the layout for RPG~\cite{yang2024mastering} is generated by GPT-4~\cite{openai2024gpt4technicalreport}. Our approach uses sketch layouts and, based on the FLUX model, achieves higher multi-instance alignment with prompts and layouts. The underlined prompts within the prompts are the sub-prompts for instances, with colors matching those in the sketch.}
    \label{fig:app-main-figures1}
\end{figure*}

\begin{figure*}[b]
    \centering
    \includegraphics[width=0.95\textwidth]{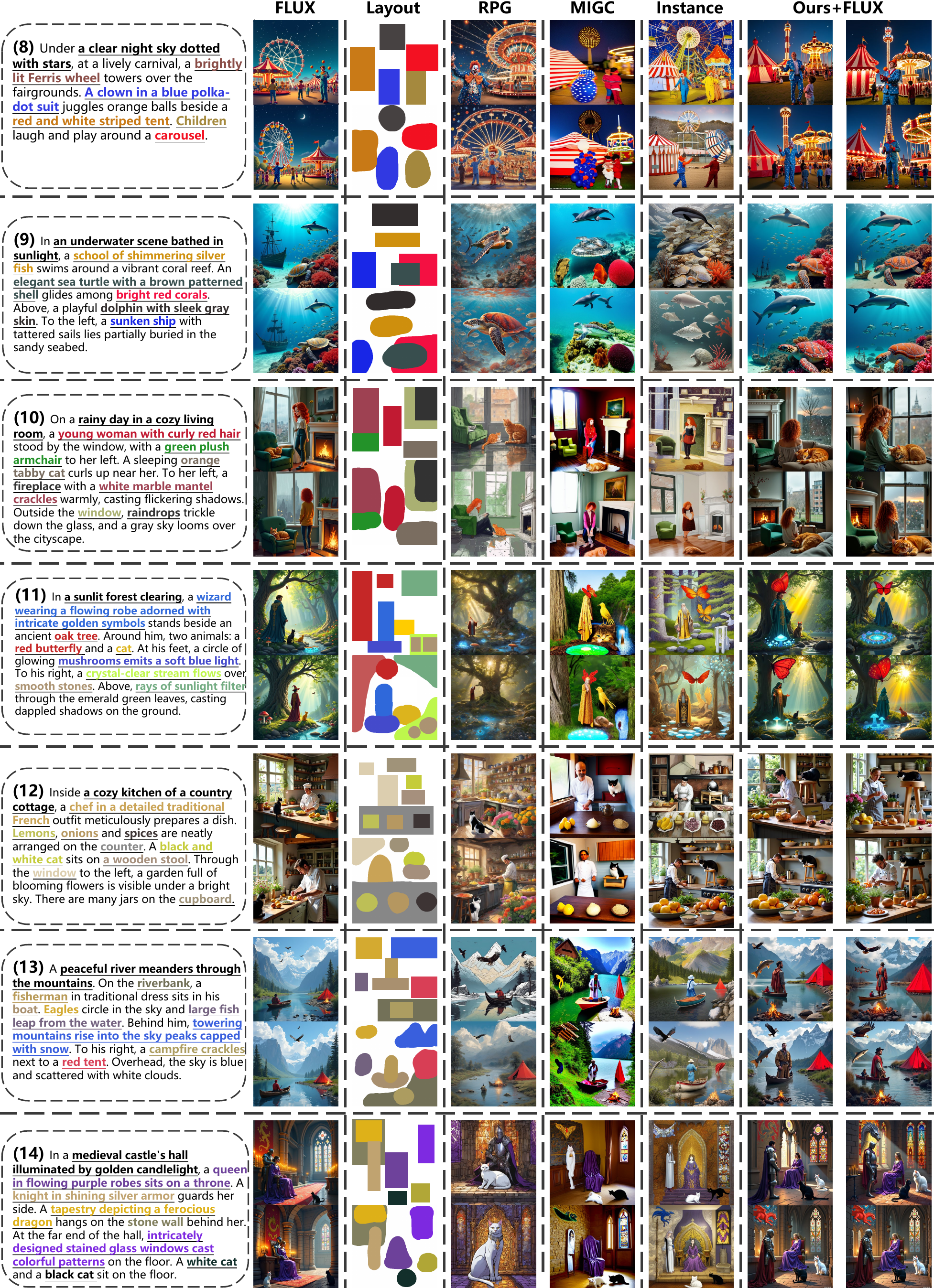}
    \caption{Comparison of 7 cases with other diffusion models, supplementing the main result (Figure 5 in the main text). Bounding box layouts are provided for MIGC~\cite{zhou2024migc} and InstanceDiffusion~\cite{wang2024instancediffusion}, while the layout for RPG~\cite{yang2024mastering} is generated by GPT-4~\cite{openai2024gpt4technicalreport}. Our approach uses sketch layouts and, based on the FLUX model, achieves higher multi-instance alignment with prompts and layouts. The underlined prompts within the prompts are the sub-prompts for instances, with colors matching those in the sketch.}
    \label{fig:app-main-figures2}
\end{figure*}

\begin{figure*}[b]
    \centering
    \includegraphics[width=1.0\textwidth]{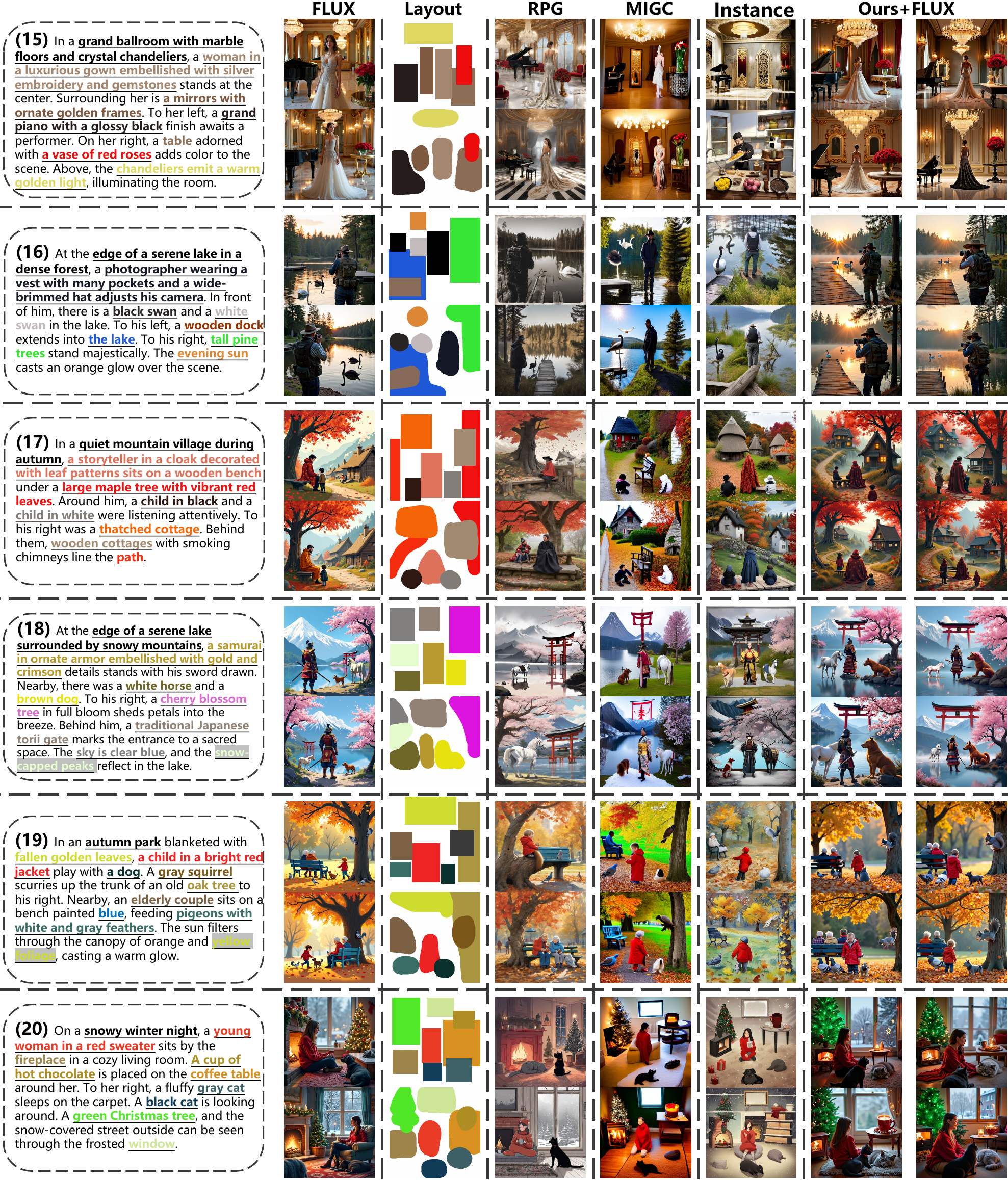}
    \caption{Comparison of 6 cases with other diffusion models, supplementing the main result (Figure 5 in the main text). Bounding box layouts are provided for MIGC~\cite{zhou2024migc} and InstanceDiffusion~\cite{wang2024instancediffusion}, while the layout for RPG~\cite{yang2024mastering} is generated by GPT-4~\cite{openai2024gpt4technicalreport}. Our approach uses sketch layouts and, based on the FLUX model, achieving higher multi-instance alignment with prompts and layouts. The underlined prompts within the prompts are the sub-prompts for instances, with colors matching those in the sketch.}
    \label{fig:app-main-figures3}
\end{figure*}

To further substantiate the findings and conclusions in Section 3.2 of the main text, we conducted three systematic token-exchange experiments. The experimental design utilized the following prompt pairs: (1) ``Red cube in a forest" and "Blue car in indoor parking", (2) ``Yellow dish in a kitchen" and "White chair in a playground", and (3) ``Orange cone in a street" and "Green cylinder on a rooftop". The comprehensive results of these token exchanges are visually documented in Figures~\ref{fig:token-exchange-three}-\ref{fig:token-exchange-two}. The model demonstrates remarkable responsiveness in early stages (by step 8), as shown by the seamless integration of color components (``Red", ``Orange", ``Yellow"), object instances (``cube", ``cone", ``dish"), and background contexts (``forest", ``street", ``kitchen"). Selective replacements of components, such as color, instance, and background, function on different layers, further confirming our prior conclusions on attention map analysis in Section 3.2. These findings enhance confidence for developing our HSLW module on the FLUX model for controllable MIS with complex prompts and specific layouts.


\end{document}